%% file: main.tex
\documentclass[lettersize,journal]{IEEEtran}

\usepackage{amsmath,amsfonts}
\usepackage{algorithmic}
\usepackage{algorithm}
\usepackage{array}
\usepackage{textcomp}
\usepackage{stfloats}
\usepackage{url}
\usepackage{verbatim}
\usepackage{graphicx}
\usepackage[caption=false,font=footnotesize,labelfont=sf,textfont=sf]{subfig}
\usepackage{cite}
\usepackage[colorlinks=true,
            linkcolor=blue,
            citecolor=blue,
            urlcolor=blue]{hyperref}
\usepackage{multirow}

\usepackage[capitalize]{cleveref}
\crefname{section}{Sec.}{Secs.}
\Crefname{section}{Section}{Sections}
\Crefname{table}{Table}{Tables}
\crefname{table}{Tab.}{Tabs.}

\usepackage{xspace}

\makeatletter
\DeclareRobustCommand\onedot{\futurelet\@let@token\@onedot}
\def\@onedot{\ifx\@let@token.\else.\null\fi\xspace}
\makeatother
\def\ie{\emph{i.e}\onedot}

\usepackage{booktabs}

\let\titleold\title
\renewcommand{\title}[1]{\titleold{#1}\newcommand{\thetitle}{#1}}
\def\maketitlesupplementary
   {
   \newpage
       \twocolumn[
        \centering
        \Large
        \textbf{\thetitle}\\
        \vspace{0.5em}Supplementary Material \\
        \vspace{1.0em}
       ] %< twocolumn
   }

\begin{document}

\title{DeepCoT: Deep Continual Transformers \\ for Real-Time Inference on Data Streams}

\author{Gin\'es Carreto Pic\'on, Peng Yuan Zhou, Qi Zhang, and Alexandros Iosifidis
\thanks{Email addresses: gcp@ece.au.dk (G. Carreto Picón), pengyuan.zhou@ece.au.dk (P. Y. Zhou), qz@ece.au.dk (Q. Zhang), alexandros.iosifidis@tuni.fi (A. Iosifidis)}
\thanks{G. Carreto Pic\'on, P. Y. Zhou, and Q. Zhang are with the Department of Electrical and Computer Engineering, Aarhus University, Denmark.}
\thanks{A. Iosifidis is with the Data Science Research Centre, Tampere University, Finland.}
\thanks{The code produced for this work is available in \protect\url{https://github.com/Atamarado/deepcot}.}
}

\maketitle

\input{sec/0_abstract}


\input{sec/1_intro}
\input{sec/2_rw}
\input{sec/3_method}

\input{sec/4_experiments}
\input{sec/5_conclusions}

\section*{Acknowledgments}
This research was supported by the PANDORA project, funded by the European Union’s Horizon Europe Framework Programme under Grant Agreement No. 101135775, and the NordForsk Nordic University Cooperation on Edge Intelligence (Grant No. 168043).

% --- Reset counters so numbering starts fresh in supplementary ---
\setcounter{figure}{0}
\setcounter{table}{0}
\setcounter{section}{0} % Sets the internal counter to 0
\setcounter{equation}{0}
\input{sec/X_suppl}

\bibliographystyle{unsrt} 
\bibliography{main}

\vfill

\end{document}

%% file: sec/0_abstract.tex
\begin{abstract}

Transformer-based models have dramatically increased their size and parameter count to tackle increasingly complex tasks. At the same time, there is a growing demand for high performance, low-latency inference on devices with limited resources. In particular, stream data inference is typically performed over a sliding temporal window, leading to highly redundant computations. While the recent Continual Transformers started addressing this issue, they can be effectively used only in shallow models, which limits their scope and generalization power. In this paper, we propose the \textbf{Deep} \textbf{Co}ntinual \textbf{T}ransformer (\textbf{DeepCoT}), a redundancy-free encoder attention mechanism that can be applied over existing deep encoder architectures with minimal changes. In our experiments over audio, video, and text streams, we show that DeepCoTs retain comparative performance to their non-continual baselines while offering a linear computational cost for all Transformer layers, which reduces up to two orders of magnitude in the running time compared to previous efficient models.
\end{abstract}

%% file: sec/1_intro.tex
\section{Introduction}
\label{sec:intro}

\noindent\IEEEPARstart{T}{ransformer} models \cite{Vaswani17transformer} have shown impressive performance for a wide range of classification and regression tasks \cite{Devlin19BERT, Dosovitskiy21ViT}. However, their size has grown significantly as new complex tasks have been targeted, resulting in slower inference speeds. This problem is especially critical in tasks where large amounts of data needs to be processed in real time, such as stream processing. 

\begin{figure}[!t]
  \centering
   \includegraphics[width=\linewidth]{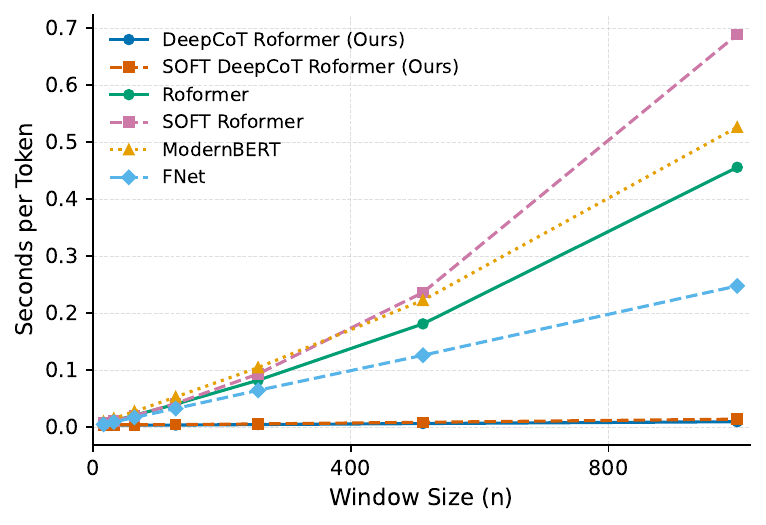}
   \caption{Average latency observed with different window sizes (batch size of 16). Latency of our DeepCoT models increases linearly with respect to the input window size ($n$), with a negligible cost increase with larger window sizes. The details of this experiment can be found in \cref{sc:runtime}.}
   \label{fig:runtime_16}
\end{figure}

In stream processing, new predictions are made by a model at specific intervals or on-demand, given new data inputs. Some data types that can have these characteristics are video inputs captured from a camera, audio inputs captured from a microphone, or text input created by a human. Transformer models typically benefit from leveraging past information together with the present data and rely on a sliding temporal window formed by the $n$ most recent data points. However, this process leads to a high level of computational redundancy, since the temporal relations between past tokens must be recomputed at every iteration. The Continual Transformers \cite{Hedegaard23cont_trans} partially addressed this issue by introducing a redundancy-free version of the Transformer encoder. However, their gains are modest when using a stack of two encoder layers, and deeper encoder models cannot include Continual Transformer layers due to the dependencies of tokens between layers. This makes Continual Transformers only suitable for problems that can be solved with one or two Transformer layers, as their efficiency advantages get washed out in deeper models.

\noindent\textbf{Contributions.}
In this work we tackle the existing limitation in the number of Continual Transformer layers by proposing the Deep Continual Transformer (DeepCoT) encoder. In a nutshell, DeepCoT uses a stack of Single Output Encoder layers \cite{Hedegaard23cont_trans} to create a redundancy-free model for stream processing tasks, leading to an encoder-model where every token looks into its present and past, as every DeepCoT layer receives one input token and outputs the attended token with linear cost, reducing the latency and hardware requirements of deep Transformer models for Continual Inference, as can be seen in \cref{fig:runtime_16}. We make the following contributions:
\begin{itemize}
    \item We propose the first deep Transformer encoder model suitable for Continual Inference, named DeepCoT.
    \item We provide a mathematical analysis of the differences between regular Transformer and DeepCoT encoder stacks.
    \item We show how existing high-performing deep Transformer models can be converted to their DeepCoT counterparts.
    \item We conduct extensive experiments on audio classification, sound event detection, online action detection, and text classification in stream data inference scenarios to compare the performance of our model against previous baselines, including both non-continual and continual adaptations of existing architectures. DeepCoTs lead to reductions of inference time up to two orders of magnitude, while retaining comparable performance to their non-continual counterparts. This is evaluated both over shallow and deeper encoders.
\end{itemize}

%% file: sec/2_rw.tex
\section{Related work}\label{sc:rw}
In this section we focus on Continual Inference Networks. We describe other relevant related work in \cref{sc:sup_rw} of the supplementary material.

\subsection{Continual Inference Networks}\label{sc:cin}
When performing inference on data streams with models that do not have inherent recurrent formulations, such as Transformers and Convolutional Neural Networks (CNNs), temporal relations among the recent data points are commonly exploited by using a temporal sliding window. This effectively transforms stream processing to successive static inference steps, \ie it slices the data stream to successive data clips each of which is introduced to the model, which provides independent responses. This process leads to redundant computations for every inference step, and further limits the model for low-latency scenarios.

Continual Inference Networks is a family of models that process stream data \textbf{without computational redundancy} while providing identical outputs to their non-continual counterparts, given the same trainable parameters \cite{Hedegaard23cont_trans}. Computational redundancy is removed by caching operations that can be reused for future inference steps. This approach has been successfully applied to temporal convolutions \cite{Hedegaard22cont_3dconv}, graph convolutions \cite{Hedegaard23cont_convgraph} and self-attention \cite{Hedegaard23cont_trans}.

In particular, the \textbf{Continual Transformers} \cite{Hedegaard23cont_trans} are Transformer encoder layers with linear cost with respect to the attention window. The Continual Nyströmformers \cite{Carreto24cont_nyst} propose two variants of the Nyströmformer \cite{Xiong21Nystromformers} that include adaptations to reduce the overall computational overhead in Continual Inference scenarios by either delaying the updates of the Nyström landmarks or using pre-computed landmarks that remain fixed during inference time. We describe the details of the Continual Transformers in \cref{sc:sup_rw} of the supplementary material. Despite some observed improvements in the Continual Nyströmformers \cite{Carreto24cont_nyst} thanks to the use of the Nyström approximation for continual layers, \textbf{it is currently not possible to stack more than two Transformer encoder layers for efficient Continual Inference}, due to the token interdependence between adjacent encoder layers. We address this limitation in this paper.

%% file: sec/3_method.tex
\section{Methodology}
\label{sc:method}

\begin{figure}[t]
  \centering
   \includegraphics[width=\linewidth]{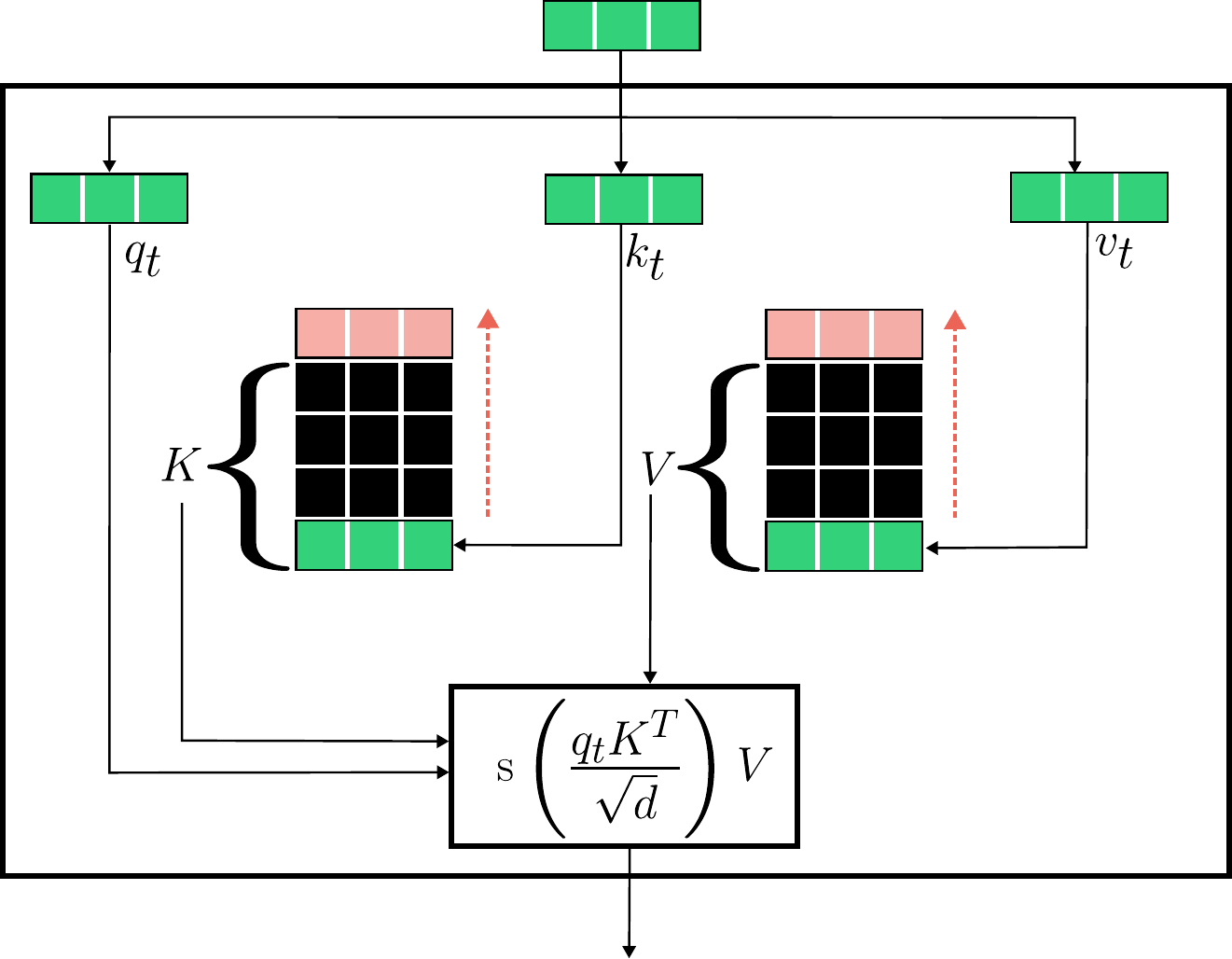}

   \caption{Overview of the attention mechanism of a DeepCoT layer. The red color represents the token that arrived $n$ time steps in the past, and has shifted out of the local attention window. The green color represents the current time step. The dashed lines indicate the temporal order in which tokens are shifted during Continual Inference.}
   \label{fig:DeepCoT_layer}
\end{figure}

\subsection{Overview}
We propose DeepCoT, a Continual Transformer encoder that performs Continual Inference over deeper models, leading to a higher predictive power compared to shallower ones. 
DeepCoT is composed by a set of $l$ stacked Transformer encoder layers, each of which uses the Single Ouput attention mechanism \cite{Hedegaard23cont_trans}. We provide a visual overview of this mechanism in \cref{fig:DeepCoT_layer}. Each encoder layer receives a single token as input with $d$ features, corresponding to the most recent temporal step, and outputs a single attended output with the last $n$ received elements (including the current token), that are kept in two memory matrices of size $(n-1) \times d$ corresponding to the Key and Value matrices. This attended token is passed through the rest of the encoder layer, and then used as input to the next layer, while the memory does not receive other updates during the inference process.

This formulation enables DeepCoT to become an efficient Transformer for Continual Inference scenarios where deeper models are needed, \ie scenarios where new data inputs arrive in a streaming fashion, with a  cost of $O(nd)$ for every layer in the model, while keeping an attention mechanism similar to that of the regular Transformer encoders \cite{Vaswani17transformer}.

\subsection{Mathematical modeling of attention differences}\label{sc:math_DeepCoT}
In this section we provide a comprehensive mathematical analysis of the differences between a set of $l$ stacked Transformer encoders, using the attention mechanism~\cite{Vaswani17transformer} (referred to as base attention), and the proposed DeepCoT attention mechanism. Each attention block includes a sliding window of $n$ elements. We assume that the two models receive the input at layer $0$ and that they have identical parameters. We use the following notation:
\begin{itemize}
    \item $q_i^l$ represents the input Query row $i$ at layer $l$ of the base attention.
    \item $KV_{i:j}^l$ represents the input matrices $KV$ (corresponding to both the Key and Value matrices), from index $i$ (exclusive) to index $j$ (inclusive) at layer $l$ of the base attention. We have decided to group together these two matrices as the temporal shift they have is identical in both cases.
    \item $\hat{q}_i^l$ and $\hat{KV}_{i:j}^l$ represent the corresponding DeepCoT input matrices. The lowercase notation $\hat{kv}_{i}^l$ represents a single pair of Key and Value vectors.
    \item $\alpha(q, KV) = \rho(q,K)V$, where $\rho(\cdot)$ represents the self-attention activation operation.
    \item $\sigma(q, KV)$ represents the function to obtain the Query, Key and Value for the next layer given the outputs from the previous layer (\ie attention, out projection, Feed-Forward and skip connections).
\end{itemize}

We analyze the relationship between arbitrary tokens, indicated by an index $i$, where $i\le t$ and $t$ corresponds to the newest token (current time step), of the two attention mechanisms. We can express the DeepCoT attention as:
\begin{equation}
    \alpha(q_t, k_t, v_t) = \text{softmax}\left(\frac{q_tK_\text{mem}^T}{\sqrt{d}}\right)V_\text{mem},
    \label{eq:DeepCoT_att}
\end{equation}
\begin{equation}
    K_\text{mem} = \begin{bmatrix} K_\text{t-n+1:t-1} \\ k_t\end{bmatrix}, \:\:\:\: V_\text{mem} = \begin{bmatrix} V_\text{t-n+1:t-1} \\ v_t\end{bmatrix}.
    \label{eq:DeepCoT_roll}
\end{equation}

To keep track of the effect of the input tokens over the subsequent layers, the following additive property needs to be satisfied:
\begin{equation}
    \sigma(q_i^l, KV^l_{a:b}) = \sigma(q_i^l, KV^l_{a:c}) + \sigma(q_i^l, KV^l_{c:b}),
    \label{eq:soft_decoupling}
\end{equation}
where $a<c<b$. Since the softmax activation does not match this requirement, we adopt the SOFT attention activation \cite{Lu21SOFT} $\rho(\cdot)$:
\begin{equation}
    \rho(q,K)=\text{exp}\left(-\frac{q \:\mathbf{\ominus}\: K}{2\sqrt{d}}\right),
    \label{eq:soft}
\end{equation}
where $q \:\mathbf{\ominus}\: K$ represents the squared Euclidean distance between each pair of tokens. It is also necessary to remove the non-linear activations in the Feed-Forward block and the Layer Normalizations \cite{Ba16Layer_Normalization} of the encoder layers, as they introduce inter-token dependencies. These modifications are done exclusively to facilitate the mathematical analysis, and during the experiments these modifications are not applied (unless otherwise specified). A detailed analysis of this can be found in \cref{sc:decouple_tokens} of the supplementary material.

With this change, the following formulas are true for the base attention:
\begin{equation}
    q^{l+1}_i, kv^{l+1}_i = \sigma(q^l_i, KV_{t-n: t}^l) = \sigma(q^l_i, KV_{t-n: i}^l) + \sigma(q^l_i, KV_{i: t}^l).
    \label{eq:base_inference}
\end{equation}
In the case of DeepCoT, this takes the form:
\begin{equation}
    \hat{q}^{l+1}_i, \hat{kv}^{l+1}_i = \sigma(\hat{q}^l_i, \hat{KV}_{i-n: i}^l) = \sigma(\hat{q}^l_i, \hat{KV}_{t-n: i}^l) + \sigma(\hat{q}^l_i, \hat{KV}_{i-n: t-n}^l).
    \label{eq:DeepCoT_inference}
\end{equation}
% In order to make \cref{eq:base_inference_2,eq:DeepCoT_inference_2} true, we need to obtain an encoder layer that makes the

\subsubsection{First layer analysis}
The decoupling introduced by \cref{eq:base_inference,eq:DeepCoT_inference} allows us to analyze the differences between the two attention formulations. The absolute difference in the outputs of the first layer $\delta_i^0$ can be expressed as:
\begin{equation}
    \delta^0_i = |\alpha({q}_i^0,{KV}^0_{i-n:t-n}) - \alpha({q}_i^0,{KV}^0_{i:t})|.
    \label{eq:first_layer_diff}
\end{equation}
The number of different tokens used for the Key and Value matrices depends on how far away the token $i$ is from $t$. As such, it is expected that farther from $t$ tokens have a corresponding higher value of $\delta_i^0$. This also means that the output corresponding to the last token $t$ is identical in both cases, because $\delta_i^0$ is zero when $i=t$. Note that if DeepCoT is formed by only one layer, its output at position $t$ would be identical to that of the base Transformer.

\subsubsection{Second layer analysis}
The differences expressed by $\delta_i^0$ get propagated to the inputs of the second layer, as shown in \cref{sc:decouple_tokens} of the supplementary material. This propagates the differences into the output differences of the second layer. As such, we can express the token differences at layer $l+1$ at time $t$ as:
\begin{equation}
    \delta^{1}_t = |\alpha(q_t^1, \Delta^0_{t-n:t}\lambda^0)|,
    \label{eq:delta_1}
\end{equation}
where $\Delta^0_{t-n:t}$ is a matrix of the different $\delta_i^0$ vectors and $\lambda^0$ is a multiplicative factor that depends on the Feed-Forward weights and the Key and Value weights.

\begin{figure}[t]
  \centering
   \includegraphics[width=\linewidth]{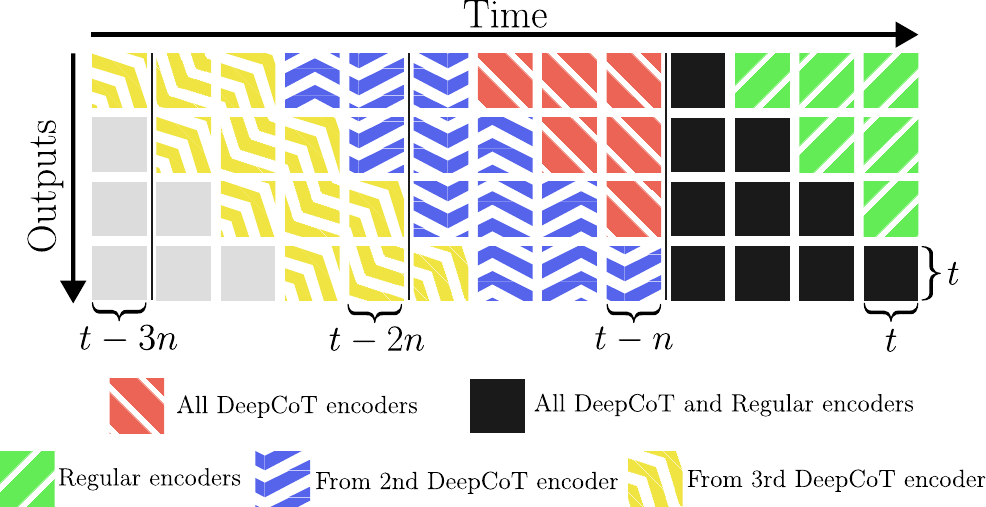}

   \caption{Comparison of the temporal receptive field of Regular encoder and DeepCoT, with $n=4$. Notice how stacking multiple DeepCoT layers extends the receptive field.}
   \label{fig:DeepCoT_sliding}
\end{figure}

In \cref{eq:DeepCoT_inference}, we can see that the matrix $\hat{KV}^1_{t-n:t}$ was obtained including elements from the input matrix at layer 0 up to $2(n-1)$ tokens in the past, as the oldest token in the second layer at the position $n-1$ uses the inputs from layer zero up to $n-1$ tokens in its past. Similarly, we can notice that the rest of the output tokens of the DeepCoT attention use input tokens up to $2(n-1) - (t - i)$ positions in the past, where $i$ is the corresponding output token. An illustration of this can be seen in \cref{fig:DeepCoT_sliding}. A similar phenomenon was also observed in Mistral \cite{Jiang23Mistral} and CNNs \cite{Luo16ReceptiveFieldCNNs}. This gives DeepCoT an increased effective temporal receptive field compared to the regular Transformers (of size $n$), as it is the case for the Continual 3DCNNs compared to the regular 3DCNNs in \cite{Hedegaard22cont_3dconv}.

\subsubsection{Subsequent layer analysis}
The DeepCoT attention works similarly for the subsequent layers. However, in this case the difference $\delta^l_t$ accounts for the differences between $\hat{q}^l_t$ and $q^l_t$:
\begin{equation}
    \delta_t^l = |\alpha(\hat{q}_t^l, \Delta^{l-1}_{t-n:t}\lambda^{l-1}) + \alpha(q_t^l, \Delta^{l-1}_{t-n:t}\lambda^{l-1})|.
    \label{eq:delta_l}
\end{equation}
The effective temporal receptive field for the computation of the last token increases by $n-1$ tokens for every additional layer. As such, the output of the $l$th DeepCoT layer at position $t$ includes up to $l(n-1)$ tokens in the past in its computation. It is also important to consider that the older tokens have essentially a smaller presence for the inference step, while the newest $n$ tokens still accumulate over 50\% of the appearances in the attention window at any layer.

\subsection{DeepCoT as KV cache Encoder}
Transformer decoders \cite{Vaswani17transformer} can be regarded as Continual Inference Networks \cite{Hedegaard23cont_trans}, since they process a stream of inputs one token at a time using a sliding window mechanism restricting the attention of every token to its past. In this case, the stream of inputs comes from the output of the model itself due to the auto-regressive nature of decoder-models \cite{Vaswani17transformer}. The KV cache technique \cite{Brown20GPT3} has been proposed as a heuristic process for Transformer decoders to store previous Key and Value tokens to avoid redundancies in computation, leading to an efficient Continual Inference mechanism.

Our DeepCoT uses a similar mechanism to achieve a linear attention cost for all its layers by storing previous Key and Value tokens in memory without updating, bounding the attention window of these tokens to their past $n$ tokens, in a similar fashion to Transformer decoders. Thus, DeepCoT defines the encoder-counterpart of the KV cache mechanism for stream processing. However, there are two key differences between these two mechanisms:
\begin{itemize}
    \item Decoder-based models are typically used for auto-regressive tasks, while encoder-based models obtain an attended token into a feature space normally used to solve classification and regressive tasks.
    \item While using the KV cache technique in decoders leads to identical results as a regular decoder, DeepCoT gives different results than a regular encoder, since the attention windows of older tokens are increased, as shown in \cref{fig:DeepCoT_sliding}.
\end{itemize}

\subsection{Converting existing models to DeepCoTs}
Even though the mathematical analysis in \cref{sc:math_DeepCoT} considers the SOFT attention activation with removed layer nonlinearities, our approach can be used to convert other existing deep Transformer models to their corresponding Continual Inference versions. Moreover, while training DeepCoTs from scratch is needed for the cases where new tasks are considered, the ability to perform transfer learning from pretrained models allows for obtaining high-performing Continual Inference models with low training time and high energy savings. We describe the process to convert existing deep Transformer models to their Continual Inference versions in \cref{sc:make_DeepCoT} of the supplementary material.

%% file: sec/4_experiments.tex
\section{Experiments}\label{sc:exp}
We evaluate performance on four different task domains: Online Action Detection (\cref{sc:oad}), Audio Classification (\cref{sc:gtzan}), Sound Event Detection (SED) (\cref{sc:sed}), and Text Classification (\cref{sc:text}). We present runtime results with longer input data in \cref{sc:runtime} and discuss the limitations and future work directions in \cref{sc:limitations}.

% We developed the implementation of DeepCoT expanding the existing \textit{continual-inference} library \cite{Hedegaard22continual_inference}.
We base our DeepCoT implementation on the \textit{continual-inference} library \cite{Hedegaard22continual_inference}. We adapt the code from the Continual Nyströmformers \cite{Carreto24cont_nyst} for the experiments over audio classification and video data, and the code for MAT-SED \cite{Cai24MAT_SED} for the SED experiments. For the text data and runtime experiments with long sequences, we used Huggingface \cite{Wolf20hf} packages to set up the experimental protocol. All training and evaluations were conducted on a NVIDIA RTX 2080 Ti GPU, unless otherwise specified.

\begin{table}
  \caption{Results for Online Action Detection on THUMOS14 \cite{Idrees17THUMOS}.
  All models use two Transformer layers. Nyström-based models use 16 landmarks. The relative runtime corresponds to the time taken to perform a forward pass through the entire validation set in a Continual Inference scenario (one token at a time). The best result is in \textbf{bold} font and the second-best result is \underline{underlined}.}
  \label{tab:OAD}
  \centering
\resizebox{\columnwidth}{!}{%
\begin{tabular}{@{}l|cccc@{}}
    \toprule
    Model & mAP K400 (\%) $\uparrow$ & mAP ANet (\%) $\uparrow$ & FLOPs (M) $\downarrow$ & Rel. Runtime ($\times$) $\uparrow$ \\
    \midrule
    OAD Transformer \cite{Wang21Oadtr} & $\mathbf{64.66\pm0.30}$ & $\mathbf{56.95\pm0.80}$ & $16.92$ & $\times 1$ \\
    Co. Transformer \cite{Hedegaard23cont_trans} & $\underline{63.93\pm0.43}$ & $56.04\pm0.27$ & $\underline{0.65}$ & $\underline{\times 10.55}$ \\
    Nyströmformer \cite{Xiong21Nystromformers} & $59.32\pm0.23$ & $50.24\pm0.14$ & $9.42$ & $\times 1.06$ \\
    Co. Nyströmformer \cite{Carreto24cont_nyst} & $59.30\pm0.39$ & $50.68\pm0.38$ & $1.43$ & $\times 0.99$ \\
    \textbf{DeepCoT (Ours)} & $63.68\pm0.43$ & $\underline{56.21\pm0.61}$ & $\mathbf{0.40}$ & $\mathbf{\times 23.65}$ \\
    \bottomrule
\end{tabular}
}
\end{table}

\subsection{Online Action Detection}\label{sc:oad}
The Online Action Detection (OAD) task objective is to detect an action as soon as possible after it begins \cite{DeGeest16OAD}. This is especially relevant for critical systems such as autonomous driving. Previous works proposed Transformers with a sliding temporal window \cite{Wang21Oadtr, Hedegaard23cont_trans, Carreto24cont_nyst} to solve this task, due to the great capacity of Transformers to solve complex tasks.

We have selected THUMOS14 \cite{Idrees17THUMOS} video dataset for evaluating our model. THUMOS14 contains 413 training videos including actions from 20 different classes, with action labels provided at the frame-level. Following previous works \cite{Wang21Oadtr, Hedegaard23cont_trans, Carreto24cont_nyst}, we use a Temporal Segment Network (TSN) \cite{Wang19temporal_segment_networks} pretrained on ActivityNet (ANet) \cite{Heilbron15ActivityNet} or Kinetics-400 (K400) \cite{Carreira17Kinetics} for all models. This process creates inputs with 64 tokens, that are fed into a Transformer model with two layers and a classification head that uses the output token corresponding to the second token to perform the action classification.

We report the mean Average Precision (mAP) on the validation set obtained both with the features obtained with the TSN trained on K400 \cite{Carreira17Kinetics} and ANet \cite{Heilbron15ActivityNet} in \cref{tab:OAD}, with five different random weight initializations. FLOPs are computed following the analysis from previous works \cite{Hedegaard23cont_trans, Carreto24cont_nyst}. We compare the performance of the baseline OAD Transformer \cite{Wang21Oadtr}, a model tailored to solve the OAD task, the Continual Transformer \cite{Hedegaard22continual_inference}, the Nyströmformer \cite{Xiong21Nystromformers}, the Continual Nyströmformer \cite{Carreto24cont_nyst}, and our proposed DeepCoT. We have removed the decoder layers used in OAD \cite{Wang21Oadtr}, leaving only the encoders used for action detection. All the models use the same architecture, following \cite{Wang21Oadtr}. Since the previous Continual Inference works only propose efficient layers up to two blocks, all architectures used for this experiments contains two layers.

The best performance is obtained by the OAD Transformer \cite{Wang21Oadtr}, followed closely by both the Continual Transformers \cite{Hedegaard23cont_trans} and DeepCoT. The Nyström-based models perform worse than the other baselines. Our DeepCoT is the most efficient model both in terms of FLOPs and runtime, with reductions of 2 orders of magnitude compared to the OAD Transformer \cite{Wang21Oadtr}. Some gains are also observed compared to the previous Continual Transformers \cite{Hedegaard23cont_trans}, where our DeepCoT model has half the runtime.

\begin{table}
  \caption{Results for Audio Classification on GTZAN \cite{Tzanetakis02GTZAN}. All models use two Transformer layers. The best result is in \textbf{bold} font and the second-best result is \underline{underlined}.}
  \label{tab:GTZAN}
  \centering
\resizebox{\columnwidth}{!}{%
\begin{tabular}{@{}l|ccc@{}}
    \toprule
    Model & Accuracy (\%) $\uparrow$ & FLOPs (K) $\downarrow$ & Rel. Runtime ($\times$) $\uparrow$ \\
    \midrule
    Transformer \cite{Vaswani17transformer} & $94.19\pm5.41$ & $11134.3$ & $\times 1$ \\
    Co. Transformer \cite{Hedegaard23cont_trans} & $\underline{94.28\pm5.44}$ & $230.7$ & $\underline{\times 1.02}$ \\
    Nyströmformer \cite{Xiong21Nystromformers} & $\mathbf{94.66\pm5.64}$ & $845.4$ & $\times 0.56$ \\
    Co. Nyströmformer \cite{Carreto24cont_nyst} & $93.53 \pm 8.53$ & $\mathbf{114.3}$ & $\times 0.71$ \\
    \textbf{DeepCoT (Ours)} & $94.19\pm5.59$ & $\underline{138.7}$ & $\mathbf{\times 37.24}$ \\
    \bottomrule
\end{tabular}
}
\end{table}

\subsection{Audio classification}\label{sc:gtzan}
Audio Classification consists of labeling an audio input into a set of categories. Transformer encoders can be used to solve this task \cite{Zaman25audiotransformers} in combination with a sliding window, enabling the possibility to apply Continual Inference improvements to it. For our experiments, we use the GTZAN Music Genre Classification dataset \cite{Tzanetakis02GTZAN}, which includes 1,000 music clips that are classified into 10 different genres. Following previous works \cite{Hedegaard23cont_trans, Carreto24cont_nyst, Zaman25audiotransformers}, we obtain the Mel Spectrograms commonly used for audio preprocessing and convert them into 120 tokens using a VGGish network.

We evaluate the performance over two Transformer encoder layers. We compare the same architectures with different attention mechanisms, including the regular Transformers \cite{Vaswani17transformer}, the Continual Transformers \cite{Hedegaard23cont_trans}, the low-rank Nyströmformer attention \cite{Xiong21Nystromformers}, the Continual Nyströmformer \cite{Carreto24cont_nyst}, and our DeepCoT. Instead of training the DeepCoT model from scratch we transfer the weights from the trained Continual Transformer model with no fine-tuning. All Nyström-based methods use 4 landmarks, with regular updates to the landmarks following \cite{Carreto24cont_nyst}.

The results can be seen in \cref{tab:GTZAN}. We average the accuracy and runtime results over a grid including five different random data splits and five different random model weight initializations. FLOPs refer to the number of operations corresponding to the attention blocks, reusing previous analysis \cite{Carreto24cont_nyst}. In the case of DeepCoT, the FLOPs are computed as the cost of a Continual Single Output attention block multiplied by the number of layers.% The runtime corresponds to the time taken by each model to fully process 100 sequences.

All the models exhibit similar accuracy from 93 to 95\%, with the best overall performance achieved by the Nyströmformers, while the rest of the models except for the Continual Nyströmformers exhibit a very similar performance. DeepCoT manages to achieve a competitive performance \textbf{without any retraining}.

When it comes to the number of FLOPs the Continual Nyströmformers \cite{Carreto24cont_nyst} achieve the lowest cost, as the combination of using the Nyström matrix with the avoidance of redundancy achieved with Continual Inference minimize the number of operations. However, our DeepCoT comes closely in the second place, as the Single Output layers convey a big reduction in the operations compared to the Retroactive layer used by the Continual Transformers \cite{Hedegaard23cont_trans}. The biggest observed gain in DeepCoT comes in terms of runtime speedup. Previous Continual models rely on the Retroactive layers, leading to an almost identical runtime compared to their corresponding non-continual layers, despite a smaller number of operations observed in the FLOPs. On the other hand, we have observed that Single Output layers lead to fast inference, as reflected in the results of DeepCoT, \textbf{with speedups of 37.24 times}. This speedup is exacerbated by the relative long window length. A reduction on runtime can also be observed when using Nyström-based layers, despite incurring into less operations.

\begin{table}
  \caption{Results for Sound Event Detection (SED) using the MAT-SED \cite{Cai24MAT_SED} architecture, composed by 10 encoder layers and three TransformerXL layers \cite{Dai19TransformerXL}. Polyphonic Sound event Detection Scores (PSDS) indicates the metric used in the DCASE2023 challenge, and SbF1 and AtF1 indicates the Segment-based and Audio tagging F1 scores for the URBAN-SED dataset \cite{Salamon17URBAN_SED}. The throughput is calculated in tokens per second (\textit{tps}), where a token is assumed to be a time step in the spectrogram input.}
  \label{tab:mat-sed}
  \centering
\resizebox{\columnwidth}{!}{%
\begin{tabular}{@{}l||cc|cc|cc@{}}
    \toprule
    \multirow{2}{*}{Model} & \multicolumn{2}{c}{DCASE2023} & \multicolumn{2}{c}{URBAN-SED}&\multicolumn{2}{c}{Efficiency metrics} \\
     & PSDS1 $\uparrow$ & PSDS2 $\uparrow$ & SbF1 $\uparrow$ & AtF1 $\uparrow$ & FLOPs (G) $\downarrow$ & Throughput (\textit{tps}) $\uparrow$ \\
    \midrule
    MAT-SED \cite{Cai24MAT_SED} & 0.0798 & 0.4185 & 0.5828 & 0.7062 & 41 & 0.532\\
    \textbf{DeepCoT MAT-SED (Ours)} & 0.0678 & 0.4610 & 0.4062 & 0.6702 & 0.284 & 8.004\\
    \bottomrule
\end{tabular}
}
\end{table}

\subsection{Sound Event Detection}\label{sc:sed}
Sound Event Detection (SED) tasks consists in detecting various predefined classes (events) in an audio clip, indicating the beginning and the end of every detected event. Transformer encoders such as MAT-SED \cite{Cai24MAT_SED} can be used to solve this task. MAT-SED is a model composed by an encoder module with 10 Transformer layers followed by a context network that includes three Transformer-XL layers \cite{Dai19TransformerXL}. We apply the DeepCoT modifications over the existing MAT-SED architecture. This is described in detail in \cref{sc:mat-sed_DeepCoT} of the supplementary material. For our experiments we use the DCASE2023 dataset \cite{Turpault19DCASE}. The tasks consists in detecting 10 different events in domestic environments in 10-second real and synthetic audio clips. Labels ware provided at the frame or clip level. Since a relevant portion of this dataset is no longer available, we also asses the performance on the URBAN-SED dataset \cite{Salamon17URBAN_SED}. This is a SED dataset composed by 10,000 synthetic audio clips including frame level annotations.

Results are presented in \cref{tab:mat-sed}. DeepCoT exhibits similar scores to that of the base model in the DCASE2023 dataset, with moderate improvements in the PSDS2 score for the DeepCoT variant. For the URBAN-SED dataset, a drop in performance is observed for the Segment-based F1 score (SbF1), which evaluates the predicted events at every second. While the original MAT-SED can leverage the information from the entire clip to make its predictions, DeepCoT attention is strictly limited to the present and past information, which makes this task significantly harder. However, the Audio tagging F1 scores (AtF1) are much closer, since this metric analyzes the quality of the predictions at the clip level, rather than at the time step level. A dramatic reduction in the number of FLOPs can be observed, which translates into an increase in throughput of $\sim15$ times, enabling real-time for SED. These results were obtained on a computer with a RTX 6000 Ada NVIDIA GPU.

\subsection{Text experiments}\label{sc:text}
We have also evaluated the performance of DeepCoT on larger models. The sequential nature of text data opens the possibility of applying sliding window-based stream processing. Such an approach can be applicable to problems such as characters being written from a keyboard \cite{Pavlopoulos20ClinicalKeyboard} or text data sent through a network. These inputs can become very large, making them suitable for Continual Inference processing.

To the best of our knowledge, no prior works have used encoder-only models to solve text tasks for stream processing scenarios. Previous works such as BERT \cite{Devlin19BERT} uses a full attention input, limiting the sequence length to a given maximum to avoid disproportionally high computation costs. By applying a Continual Inference, we can obtain significant computational cost reductions.

We perform our evaluations on the GLUE \cite{Wang19GLUE} benchmark. We use seven different tasks suitable for encoder-only models, such as single-sentence classification, pair sentence similarity computation, and semantic analysis tasks. We evaluate each task using three different window sizes, \ie equal to $\times 0.5$, $\times 1$, and $\times 2$ the average sequence length in the validation set. The specific window sizes can be seen in \cref{tab:GLUE_complete}.
The process used for adapting the dataset to a Continual Inference setting and evaluating the efficacy of the different models both during fine-tuning and evaluation time can be found in \cref{sc:text_finetuning} of the supplementary material. Training of all models was conducted on RTX 6000 Ada NVIDIA GPUs due to the high memory cost required during training, while the running evaluations were conducted on a RTX 2080 Ti NVIDIA GPU.

We adapt the Roformer model \cite{Su24Roformer} for the design of our proposed \textbf{DeepCoT Roformer}. The Roformer uses a BERT-like architecture \cite{Devlin19BERT} with 12 Transformer layers and Rotary Position Embedding (RoPE). Rotation-based embeddings are circular by definition, making them compatible with Transformers in Continual Inference scenarios \cite{Hedegaard23cont_trans}. We also evaluate the performance of the models using the SOFT activation function \cite{Lu21SOFT} both for the Roformer \cite{Su24Roformer}and the DeepCoT Roformer, following the mathematical analysis presented in \cref{sc:math_DeepCoT}. As such, we also replace all the Layer Normalization layers in the Transformers \cite{Ba16Layer_Normalization} with ReZero layers \cite{Bachlechner21ReZero} with a constant multiplicative factor $\alpha = 1/l$, where $l=12$ is the number of layers. These changes align the models with our mathematical analysis. All of the Roformer-based models in our text experiments use the original Roformer pretrained weights as the starting point for fine-tuning.

In our evaluations we also include two additional baselines. ModernBERT \cite{Warner25ModernBERT} is a revisited version of the BERT architecture \cite{Devlin19BERT} with various changes to improve the performance of the model, including the use of RoPE \cite{Su24Roformer}. FNet \cite{Lee22FNet} is an efficient text processing architecture that replaces the attention mechanism with Fourier transforms, obtaining a model with $O(n \text{log}(n))$ computational cost. Both models have available pretrained checkpoints on Hugging Face \cite{Wolf20hf} that we use for fine-tuning, following the same procedure described in \cref{sc:text_finetuning} of the supplementary material.
\begin{table*}
    \caption{Results on the GLUE benchmark \cite{Wang19GLUE}. The numbers in parentheses indicate the window size used for each experiment. The first number in every cell refers to the performance, and the second number to the throughput of the model computed in tokens per second. Additional aspects of this experiment can be found in \cref{sc:text_finetuning} of the supplementary material. The best result is in \textbf{bold} font and the second-best result is \underline{underlined}.}
    \label{tab:GLUE_complete}
\resizebox{\linewidth}{!}{%
\begin{tabular}{@{}l|c||ccccccc@{}}
    \toprule
    Metric & - & F1 & F1 & F1 & MAE score & F1 & Acc. & F1 \\
    \midrule
    $\times 0.5$ & \textbf{Average} & CoLA (6) & SST-2 (12) & MRPC (26) & STS-B (15) & QQP (15) & MNLI (19) & QNLI (25) \\
    \midrule
    SOFT Roformer & 69.01 / 57.0 & 79.42 / 73.7 & 72.02 / 58.0 & \underline{80.53} / 43.4 & 69.82 / 64.9 & 72.35 / 58.4 & 56.59 / 53.0 & 52.34 / 47.3 \\
    \textbf{DeepCoT SOFT Roformer (Ours)} & \underline{74.71} / 73.7 & \underline{81.40} / 80.8 & \underline{83.67} / 63.6 & 80.00 / \underline{81.1} & 72.27 / 81.0 & 76.08 / 82.4 & \underline{66.40} / 63.4 & \underline{63.14} / 63.6 \\
    ModernBERT \cite{Warner25ModernBERT} & 72.66 / 49.4 & 75.79 / 50.5 & 81.48 / 52.3 & 79.80 / 37.0 & \underline{77.27} / 52.0 & \underline{77.20} / 57.9 & 57.47 / 54.6 & 59.58 / 41.7 \\
    FNet \cite{Lee22FNet} & 67.71 / \textbf{112.3} & \textbf{81.75} / \textbf{158.3} & 77.86 / \textbf{144.1} & 76.35 / 71.8 & 74.47 / \textbf{122.9} & 74.17 / \textbf{121.6} & 31.82 / \underline{93.9} & 57.54 / \underline{73.5} \\
    Roformer \cite{Su24Roformer} & 69.52 / 80.1 & 79.58 / \underline{94.1} & 79.42 / \underline{79.2} & 80.51 / 59.0 & 76.57 / \underline{94.3} & 76.07 / 95.6 & 35.33 / 77.7 & 59.18 / 60.9 \\
    \textbf{DeepCoT Roformer (Ours)} & \textbf{80.81} / \underline{86.4} & 80.97 / 77.3 & \textbf{87.93} / 77.7 & \textbf{80.90} / \textbf{99.6} & \textbf{79.84} / 76.1 & \textbf{82.80} / \underline{98.5} & \textbf{74.76} / \textbf{97.1} & \textbf{78.44} / \textbf{78.2} \\
    \midrule
    $\times 1$ & \textbf{Average} & CoLA (12) & SST-2 (24) & MRPC (52) & STS-B (30) & QQP (30) & MNLI (38) & QNLI (50) \\
    \midrule
    SOFT Roformer & 72.59 / 39.3 & 79.13 / 65.9 & 79.91 / 44.8 & 76.66 / 27.3 & 72.79 / 39.6 & 78.28 / 34.6 & 63.53 / 35.5 & 57.80 / 27.5 \\
    \textbf{DeepCoT SOFT Roformer (Ours)} & 76.83 / \underline{74.9} & \underline{81.82} / 78.8 & 82.70 / 63.1 & 81.19 / \textbf{82.8} & 73.52 / \underline{78.6} & 79.46 / 63.3 & 70.33 / \underline{76.7} & 68.76 / \textbf{80.9} \\
    ModernBERT \cite{Warner25ModernBERT} & 83.13 / 32.7 & 77.21 / 39.7 & \underline{89.39} / 41.1 & 84.69 / 21.9 & 82.28 / 35.2 & \textbf{87.17} / 38.9 & \textbf{80.23} / 29.2 & \underline{80.94} / 23.2 \\
    FNet \cite{Lee22FNet} & 77.79 / 70.1 & 81.64 / \textbf{146.5} & 84.68 / \textbf{80.5} & 77.11 / 41.4 & 81.67 / 66.3 & 81.42 / \underline{66.4} & 65.58 / 50.5 & 72.41 / 39.2 \\
    Roformer \cite{Su24Roformer} & \underline{83.39} / 53.6 & \textbf{84.64} / \underline{92.1} & \textbf{89.59} / 64.9 & \underline{85.01} / 34.5 & \underline{83.22} / 54.1 & \underline{85.93} / 54.9 & 75.74 / 42.7 & 79.61 / 32.2 \\
    \textbf{DeepCoT Roformer (Ours)} & \textbf{84.80} / \textbf{84.4} & 80.91 / 78.0 & 89.23 / \underline{77.8} & \textbf{85.95} / \underline{80.0} & \textbf{86.48} / \textbf{99.1} & 85.79 / \textbf{76.8} & \underline{79.04} / \textbf{100.1} & \textbf{86.18} / \underline{78.8} \\
    \midrule
    $\times 2$ & \textbf{Average} & CoLA (24) & SST-2 (48) & MRPC (104) & STS-B (60) & QQP (60) & MNLI (76) & QNLI (100) \\
    \midrule
    SOFT Roformer & 76.63 / 28.5 & 80.40 / 57.0 & 83.08 / 44.8 & 81.89 / - & 73.75 / 21.4 & 80.56 / 21.4 & 71.02 / 14.5 & 65.69 / 11.9 \\
    \textbf{DeepCoT SOFT Roformer (Ours)} & 77.68 / \underline{66.3} & 81.04 / 82.5 & 82.98 / \underline{78.9} & 82.19 / - & 74.12 / \underline{53.7} & 80.33 / \underline{62.8} & 72.21 / \underline{55.8} & 70.88 / \underline{64.4} \\
    ModernBERT \cite{Warner25ModernBERT} & \textbf{88.15} / 22.6 & 78.98 / 39.3 & \textbf{93.08} / 25.5 & \underline{87.93} / - & \textbf{89.11} / 22.7 & \textbf{88.26} / 20.7 & \textbf{87.21} / 14.9 & \textbf{92.48} / 12.6 \\
    FNet \cite{Lee22FNet} & 82.60 / 45.8 & 80.54 / \underline{96.2} & 89.11 / 62.2 & 83.14 / - & 84.31 / 35.6 & 84.03 / 34.2 & 73.87 / 26.4 & 83.20 / 20.3 \\
    Roformer \cite{Su24Roformer} & \underline{87.72} / 40.4 & \textbf{84.53} / 72.8 & \underline{91.83} / 71.1 & \textbf{88.47} / - & \underline{87.96} / 32.6 & \underline{87.61} / 27.6 & \underline{83.58} / 21.5 & \underline{90.03} / 17.1 \\
    \textbf{DeepCoT Roformer (Ours)} & 85.18 / \textbf{95.3} & \underline{81.35} / \textbf{97.2} & 89.33 / \textbf{82.4} & 84.31 / - & 87.54 / \textbf{92.6} & 85.93 / \textbf{98.9} & 80.93 / \textbf{98.9} & 86.88 / \textbf{101.5} \\
    \bottomrule
\end{tabular}
}
\end{table*}

The detailed results obtained for GLUE can be found in \cref{tab:GLUE_complete}. We report the F1 score for all tasks, except for MNLI where we use the accuracy, and STS-B where we use the MAE (Mean Average Error) score. The MAE score is computed as $1 - \text{MAE}/5$. There is a noticeable difference in performance depending on the window length. On the smallest attention windows ($\times 0.5$), the DeepCoT-based models achieve the highest average performance. Due to the increased effective temporal receptive field, these models have access to data that have already left the attention window, allowing them to achieve higher performance than the other models. This difference gets reduced as the window size increases. We believe that this happens because the length of input sequences bounds the effective temporal receptive field of DeepCoTs (thus receiving less information than they are capable to process), while the baseline models get access to more information with their regular receptive fields, which become comparable to the length of the input sequences. Both ModernBERT \cite{Warner25ModernBERT} and Roformer \cite{Su24Roformer} achieve a slightly higher performance than the DeepCoT model with attention windows of size $\times 2$, since most of the data is available at the current window, and these models make use of the complete, bi-directional attention. However, DeepCoT keeps good performance with drops of only 2.5 points compared to the Roformer \cite{Su24Roformer}. This shows that DeepCoT offers competitive performance to the baselines even for inputs with shorter sequence lengths.
%\textbf{DeepCoT adapts to a high degree its lack of bi-directional attention}, in combination with their increased effective attention receptive field.
The SOFT-based methods fall behind in terms of performance. However, the DeepCoT SOFT version offers in general higher performance than the SOFT Roformer.

In terms of throughput, FNet \cite{Lee22FNet} offers lower latency for windows of less than 25 tokens. However, their throughput gets reduced significantly with higher windows, due to its asymptotic cost of $O(n\text{log}(n))$. Moreover their performance is in general inferior to all the other non-SOFT-based models. A slight increase in throughput is observed for DeepCoT with larger window sizes for this experiment. We believe that this is caused by some specific low-level optimizations that may benefit our model, as discussed in \cref{sc:runtime}. As such, DeepCoT offers throughput values between 75 and 100 tokens per second (tps), offering half the latency for large window sizes compared to the other models.

\subsection{Inference speed for long sequence data}\label{sc:runtime}
We also conducted runtime experiments over longer data sequences in order to evaluate the inference speed over a realistic Continual Inference scenario. In particular, we evaluate the evolution of the speed with increasing window sizes, following previous works \cite{Xiong21Nystromformers, Wang20Linformer}. We have decided to adapt the MNLI text dataset \cite{williams18MNLI} by stitching all text inputs in the evaluation sets into $b$ groups, where $b$ is the target batch size. We also add a special separation token to indicate the separation between inputs, following \cite{Faw25Stitching_time}. We then track the latency of the models described in \cref{sc:text} to produce the outputs for increasing window lengths, until we hit an out-of-memory error for any of the models. All of the models in these experiments have $d=768$ dimensions per token. We conducted these experiments on a desktop computer with an Intel I9-13900K Processor and a single NVIDIA RTX 2080 Ti GPU on Ubuntu 22.04.

The results of token latency and throughput during inference can be seen in \cref{fig:runtime_16} for batch size of 16. Additional results can be found in \cref{sc:runtime_supplementary} of the supplementary material. The first observation is that the models follow the expected computational costs in both experiments: $O(n^2)$ for Roformer \cite{Su24Roformer} and ModernBERT \cite{Warner25ModernBERT} models, $O(n \text{log} (n))$ for the FNet \cite{Lee22FNet} models and $O(n)$ for our DeepCoT models. In particular, both DeepCoT models show a small degradation in inference speed for longer sequences, which makes our model ideal for such scenarios. In particular, the tokens per second observed for DeepCoT with batch size of 16 for a window size of 64 is $303.49$, while for a size of 1000 is $106.75$. A small increase in throughput for DeepCoT is also observed for window sizes around 32 elements, compared to smaller windows. This can be related to some specific low-level optimizations that benefit DeepCoT. The use of the SOFT activation \cite{Lu21SOFT} leads to a small increase in the latency, compared to their non-SOFT counterparts. These results show that adopting the DeepCoT variant of a deep Transformer architecture for Continual Inference can effectively lead to negligible cost increase as the window size ($n$) increases.

\subsection{Limitations and future work}
\label{sc:limitations}

We observed through the different experiments how the DeepCoT behaves empirically over various existing architectures. DeepCoT offers a low-latency model that effectively retains the performance in most cases compared to its non-continual counterparts. These gains are focused on stream processing scenarios, and are not suitable for static inference settings where entire data items are introduced to the model, such as image classification. It is also worth noting that all speedups reported for DeepCoTs have been obtained by using standard PyTorch implementations. Optimized implementations and the combination of DeepCoT with other efficient Transformers could lead to further speedups.

We also observed that if most or all of the necessary data is located within the current attention window, the regular encoder-based models show a slightly higher performance. This can be critical for certain applications. In these cases, a longer window size can be applied to leverage older information to get an efficient model with the same performance. A combination of DeepCoT and regular encoder layers can also be used to improve the overall performance.

The observed runtime and FLOPs gains are dependent on the attention window size $n$. This difference can be, for example, observed in the results over text data with various window sizes. As such, DeepCoT offers greater throughput gains for larger values of $n$. The memory cost of DeepCoT can in some situations be higher to that of regular Transformers \cite{Vaswani17transformer}. While regular Transformers need to store the $QK^T$ product (of size $n \times n$), it is necessary to keep at all times the Key and Value memory matrices ($(n-1) \times d$ for every layer $l$). As such, the memory cost of Transformers is $O(n^2)$, while that of DeepCoT is $O(ndl)$.

%% file: sec/5_conclusions.tex
\section{Conclusions}\label{sc:concl}

In this work we propose DeepCoT, a Deep Continual Transformer encoder that has a linear computational cost with respect to the input sequence length $n$ for stream processing. We provide a comprehensive analysis of the differences between the attention mechanism in DeepCoT and the corresponding non-continual Transformer encoder. We show how existing deep Transformer encoder models can be converted to their corresponding DeepCoT versions allowing for leveraging pretrained model parameters. We evaluate DeepCoT on audio, video and text data domains, showing big gains in Continual Inference efficiency both in terms of computational cost and FLOPs while retaining the performance of the corresponding baselines. All these features make DeepCoT a very strong model to consider for stream processing scenarios where fast, high-performing Transformers are required and open new possibilities for adopting recent advances brought by deep Transformers into the Real-time Inference domain.

%% file: sec/X_suppl.tex
\clearpage

\maketitlesupplementary
\section{Extended related work}\label{sc:sup_rw}

\subsection{Efficient Transformer models}\label{sc:efficient_transformers}
There is a plethora of methods that have effectively reduced the quadratic computational cost $O(n^2)$ with respect to the number of input tokens $n$ of the regular Transformers \cite{Vaswani17transformer, Tay22EfficientTransformersSurvey} leading to efficient data processing in the so-called \textit{static inference} setting, \ie when a model makes predictions for data inputs that are entirely introduced to the model, e.g., as in image or video-clip classification.

The most common way to reduce the overall cost is to \textbf{reduce the number of operations}. LTP \cite{Kim22LTP} prunes the tokens with low attention result in the first Transformer layers, so the cost of subsequent layers is smaller. The Linformer \cite{Wang20Linformer} projects the Key and Value matrices into a smaller set of tokens using Singular Value Decomposition. The Nyströmformer \cite{Xiong21Nystromformers} uses the Nyström technique to approximate the multiplication between the Query and Key matrices. SOFT \cite{Lu21SOFT} argues that the softmax activation limits the ability to reduce the cost of models and combines the Nyström approximation with an activation function based on the distance between every pair of tokens in the Query and Key matrices as an alternative to softmax. Other models such as FNet \cite{Lee22FNet} replace the attention mechanism with a Fourier Transform over the input tokens. Some methods combine multiple improvements to the attention together. ModernBERT \cite{Warner25ModernBERT} uses the Rotational Positional Embeddings (RoPE) \cite{Su24Roformer} and FlashAttention \cite{Dao22FlashAttention, Shah24FlashAttention3}, a hardware-optimized version of the Transformer, which leads to some reduction in the model overhead.

Some works apply different techniques to achieve a \textbf{linear attention cost for Transformer decoders} \cite{Sun25EfficientLLMSurvey}. LoLA \cite{McDermott25LoLA} proposes a linear attention mechanism that combines three token types to retain the information from past tokens: a full rank local token window, a low-rank token set to represent past inputs, and sparse tokens that are particularly difficult to represent in the low-rank token set. Another approach for obtaining a linear attention replaces the softmax activation with an activation that allows decoupling, as described in \cite{Katharopoulos20LinearAttention}. This essentially enables linear decoders with similar properties to a regular Transformer decoder.

Other works proposed ways to \textbf{efficiently process a sliding window input using decoders}. The Sparse Transformers \cite{Child19SparseTransformers} compute the full attention window in their neighborhood, with sparse attention computations to other parts of the input image. The Longformer \cite{Beltagy19Longformer} uses a similar idea for encoder-only models. These works inspired Mistral \cite{Jiang23Mistral}, a method that applies a sliding window into their decoder-based model. 
Reducing the effective attention window can also be used to focus the attention into individual parts of the input. This is the case of the Swin Transformer \cite{Liu22VideoSwin}. Some works iterated over this idea, such as the DW-ViT \cite{Ren22DWViT}, which uses different window sizes to compute the local attention, in a similar way to some InceptionNets \cite{Szegedy16Inception}. The VSA \cite{Zhang22VariedWindowViT} applies overlapping windows the location and size of which are decided based on the input data, aiming to find the most relevant windows for the specific task.

Overall, many works target at reducing the quadratic cost of the regular attention mechanism. However, there are few works that tackle this limitation in the context of stream data inference \cite{Tay22EfficientTransformersSurvey}.

\subsection{Large Transformer models}\label{sc:large_transformers}
Multiple models stack multiple Transformer layers \cite{Vaswani17transformer} with the goal to obtain models with bigger capacity that solve more complex tasks \cite{Raiaan24LLMReview}. Consequently, recent models have grown in depth and the number of parameters has been considerably increased. Based on their use of encoders and decoders, we can divide the existing Transformer-based models into two main categories:

\noindent \textbf{Encoder-based models} extract representations, which are typically used to solve classification or regression tasks. One of the most popular encoder-only models in the language domain is BERT \cite{Devlin19BERT}, a large model pretrained on a big corpus of text data for a masked token in a self-supervised model. BERT is then fine-tuned to solve various tasks. Some subsequent works have improved the performance following different research directions, such as RoBERTa \cite{Zhuang21RoBERTa}, which optimizes some parameters of BERT, or RoFormer \cite{Su24Roformer}, which proposes a new positional encoding based on rotation matrices.

Similar advances can be observed in the vision domain with the introduction of the Vision Transformers (ViT) \cite{Dosovitskiy21ViT}, which adapts BERT for the Image Classification domain. ViT was later used as a pretrained checkpoint to create the Video Vision Transformers (ViViTs) \cite{Arnab21ViViT}, which expands the attention into the temporal domain that is presented in video inputs. Different variants of ViViTs were proposed, and the model that separates spatial and temporal attention offered the best trade-off between performance and computational cost. There have been other works that analyzed different ways to perform spatio-temporal attention \cite{Chen25VideoTrans, Bertasius21TimeSformer}.

\noindent \textbf{Decoder-based models} are mostly used for generative and forecasting tasks. In particular, many of the most successful models are decoder-only foundation models, such as GPT \cite{Brown20GPT3} for language generation, and TimesFM \cite{Das24TimesFM} for time-series forecasting. These models have also shown good performance in few-shot learning \cite{Brown20GPT3, Faw25FewShotTimesFM}. These foundation models require a big number of computations. As such, different techniques have been proposed, such as efficient KV caches \cite{Shi24KV_optimal, Li25KV_Acceleration} or sparse attention \cite{Xiao25MiniCPM}. Some works have also included decoders to enhance encoder models, such as DeBERTa \cite{He21Deberta, He23DeBERTaV3}. However, despite the observed improvements in efficiency and performance, encoder layers still outperform decoder models in some tasks, such as word meaning comprehension \cite{Qorib24EncodervsDecoder}, or offer a less costly alternative \cite{Obeidat25EncoderBiomedical}.

\subsection{Continual Transformers}\label{sc:cont_trans}
The Continual Transformers \cite{Hedegaard23cont_trans} make use of \textbf{two different Continual Inference attention blocks}. The Retroactive Attention performs inference steps by being retroactively updated with the information from the new token to the older tokens of the attention window stored in memory, and by rolling the new token into the memory matrices. The Single Output Attention performs inference steps without updating the previous elements. This is useful when only the last output of the token is required. In this case, the attention block simply performs the attention update corresponding to the last token, and rolls the corresponding elements of that token for future inference steps, without any retroactive updates.

Both of these attention blocks produce identical outputs to those of the regular attention \cite{Vaswani17transformer}. However, this does not hold for deeper Continual Transformers, because the first layer outputs different tokens than the previous iterations. \textbf{This renders the memory matrices of subsequent continual layers obsolete}, forcing a complete attention step identical to that of the regular attention. The Single Output Attention can be used in the last Transformer layer, because we only require the output of the newest token. As such, the authors proposed the following architecture for multi-layer Continual Transformers:
\begin{itemize}
    \item For one-layer encoder models, the Single Output Attention is used.
    \item For two-layer encoder models, the Retroactive Attention is used in the first Transformer layer, and the Single Output Attention is used in the last layer. This Retroactive Attention is not fully continual as it needs to remove the effect of the old token, \textbf{degrading significantly the speedup gains} compared to the Single Output Attention.
    \item For larger encoder models, the Retroactive Attention is used in the first layer and the Single Output Attention is used in the last layer. All intermediate blocks are non-continual layers, essentially leading to a non-Continual Inference architecture, as the attention windows need to be processed from scratch for all the intermediate layers of the architecture.
\end{itemize}

This limits the range of applications where the Continual Transformers can be used to shallow Transformer models such as \cite{Slama25CoGraphTransformers}. The Continual Nyströmformers \cite{Carreto24cont_nyst} make use of the Nyström matrix approximation scheme of Nyströmformers \cite{Xiong21Nystromformers} to reduce the computational cost of calculating the attention. This leads to efficient Retroactive and Single Output Nyström-based Transformers suitable for redundant-free Continual Inference for shallow models. For building deeper models, the same approach as the one described above for the Continual Transformers is followed, \ie the first and last layers of the architecture are continual Nyströmformer layers and the remaining layers are non-continual Nyströmformer layers. This leads to a lower computational cost compared to that of the Continual Transformers, however, \textbf{it essentially corresponds to a non-Continual Inference architecture for deeper models}.

\section{Decoupling the token operations in Transformers}\label{sc:decouple_tokens}
Here, we describe a decoupled Transformer that matches the requirement of Eq. (3) in Sec. 3 of the paper. We first consider the softmax attention activation applied row-wise in the Transformer:
\begin{equation}
    \text{softmax}(x_i) = \frac{\text{exp}(x_i)}{\sum_{j=0}^{j=n} \text{exp}(x_j)},
    \label{eq:softmax}
\end{equation}
where $n$ is the attention window size. The normalization in the denominator introduces inter-token dependencies among all the elements in a row, since a change to any element before the softmax activation triggers further changes to every other element in that row \cite{Katharopoulos20LinearAttention}. Choosing an activation function suited for linearization, such as SOFT \cite{Lu21SOFT}, solves this issue. SOFT uses pair-wise squared Euclidean distances between the Query and Key vectors to compute the attention, allowing for the decoupling.

Layer Normalization \cite{Ba16Layer_Normalization} also introduces inter-token dependencies, as every token is normalized by using all the other tokens. Layer Normalization is used in the Transformer layers after the merging of the main and skip connection branches for the attention and Feed-Forward blocks. As such, we replace both normalization layers with ReZero layers \cite{Bachlechner21ReZero}, in order to stabilize the gradients during fine-tuning. ReZero applies a small multiplicative factor to the main branch before the addition with the skip connection. This also ensures the property of linearity.

Finally, the non-linear activation used in the Feed-Forward block also introduces inter-token dependencies, since some tokens' outputs collapse in non-linear activations. As such, we remove the activation of such block.

These changes are made to enable better token traceability due to the interdependence created by non-linear activations and layer normalizations. During the experiments, we apply the DeepCoT modifications over the regular Transformers, unless otherwise mentioned.

\subsection{Forward propagation difference in multi-layer Transformers}\label{sc:forwardTransformerLayer}
In deep Transformers, the differences between different attentions $\delta_i^l$ at a layer $l$ get propagated into the following layer. In particular, if we consider the changes described in \cref{sc:decouple_tokens}, and we follow sequentially the operations of Transformers between the output of the activation block and the input of the activation block of the following layer we have:
\begin{enumerate}
    \item A linear projection $W^l_O$ corresponding to the output of the attention block.
    \item A linear projection corresponding to the Feed-Forward block $W^l_{FF}$, and the hyperparameter $\beta$ from the ReZero layer \cite{Bachlechner21ReZero}.
    \item The skip connection of the Feed-Forward block, which propagates the differences.
    \item The projection to obtain the Key and Value matrices of the following layer, that use $W^{l+1}_K$ and $W^{l+1}_V$, respectively.
\end{enumerate}

With these, we can model the multiplicative component of the difference $\lambda^l$ present in Eqs. (10) and (11) of the paper as:
\begin{equation}
    \lambda^l = \{\mu^l W^{l+1}_K, \mu^lW_V^{l+1}\},
\end{equation}
where $\mu^l=W^l_O (1 + W^l_{FF}\beta)$, with $1$ corresponding to the contribution of the skip connection over the Feed-Forward block. The skip connection of the attention block does not have an impact in the differences in this case, as this is already considered in $\delta^l_i$. $\lambda^l$ is an operator that encapsulates the necessary operations to obtain the Key and Value matrices.

\section{How to convert non-Continual deep Transformer models to their DeepCoT versions}\label{sc:make_DeepCoT}
DeepCoT can be applied over any existing encoder-based Transformer architecture that needs to be used in a Continual Inference setting. This Section provides some indications on how to transform existing Transformer architectures to their corresponding DeepCoT versions.

The first step consists on identifying the Transformer layers that operate over the temporal dimension. Some models include encoders that are applied only over non-temporal dimensions, such as a spatial ViT \cite{Dosovitskiy21ViT}. These do not require any changes. For all temporal Transformers, it is important to change the positional embedding into a circular positional embedding; this means, a positional embedding where its last and first position are semantically related. This is necessary as the model should be able to handle sequences of any arbitrary length over time. Some examples of circular positional embedding are the Recycling Positional Embedding \cite{Hedegaard23cont_trans} and RoPE \cite{Su24Roformer}. The temporal Transformers can then be replaced with DeepCoT layers. Skip connections can be left unchanged, since every DeepCoT layer outputs the same token positions it has received as input. In the case of BERT-like models such as Roformer, all layers compute attention between different tokens. As such, all Transformer layers are converted into DeepCoT layers. Roformer uses RoPE as its Positional Embedding. This is already a circular Embedding, and consequently it does not need to be changed.

In the case of Roformer, we assumed that one-token is introduced to the model at every inference step. However, there can be cases where $m$ tokens are introduced to the model at the same time, where $m>1$. This can, for instance, be the case for processing video streams in a continual manner with the tokens corresponding to each video frame being introduced to the model at each inference step, or processing text streams with the tokens from multiple words or a complex word introduced to the model. There are multiple solutions for these situations: 1) the single-token Continual Inference can be performed one-token at a time, 2) a Transformer encoder or another layer can be used before the existing encoder to convert the $m$ tokens into one, or 3) an $m$-output DeepCoT block can be used instead. That is, if $m$ temporal tokens ($1<m<n$) are introduced to the model and are updated at every time step, it is necessary to convert the corresponding layers into $m$-output Continual Inference Transformers. This variant of a DeepCoT layer receives $m$ tokens as input and outputs $m$ tokens as the output. It uses a matrix $Q^t$ containing the $m$ tokens corresponding to the current time step, and rolls of $m$ positions at every time step are performed to accommodate all the new information. The output of this module corresponds to the sum of the outputs of the unidirectional attention between the new and previous tokens in the memory and the full attention between the new tokens.

The use of [CLS] tokens \cite{Devlin19BERT} is generally not recommended, as they introduce some additional overhead that may reduce the efficiency of DeepCoT. Moreover, the use of [CLS] tokens in deeper Continual Transformers does not improve performance \cite{Hedegaard23cont_trans}. However, there may be some tasks where [CLS] tokens are necessary or beneficial. In those cases, they should be regarded as a new temporal token and be updated accordingly. These tokens can then either be discarded or introduced into the Key and Value memory matrices. The DeepCoT Roformers do not make use of [CLS] tokens, and we instead perform inference from the newest token.

In some situations, the models can benefit from performing fine-tuning with these changes applied. This depends on the specific task and architecture used. In any case, DeepCoT models can leverage the information learned from previous non-continual trainings to reduce the training time by initializing their parameters with the parameters of the corresponding pretrained non-Continual deep Transformer models. For our DeepCoT Roformers, we reuse the pretraining done on the regular Roformers, and we only perform fine-tuning.

\begin{figure*}[t]
  \centering
   \includegraphics[width=\linewidth]{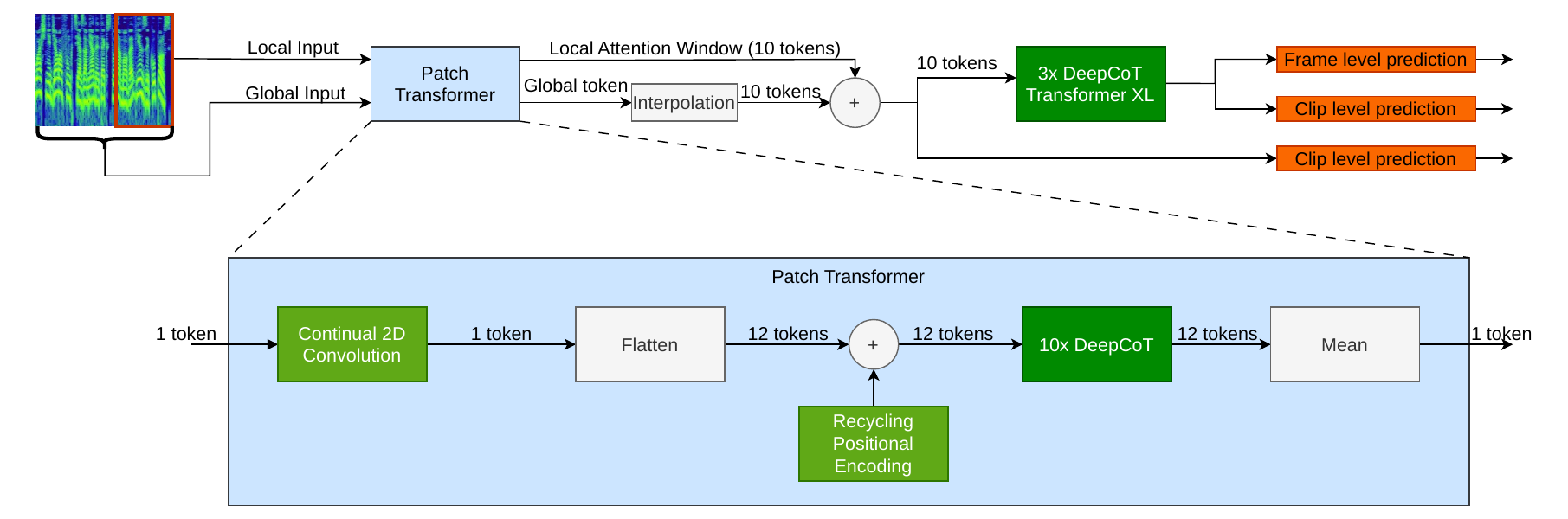}

   \caption{Overview of the DeepCoT MAT-SED architecture. Green is used to highlight the modified components with respect to the original MAT-SED architecture \cite{Cai24MAT_SED}.}
   \label{fig:DeepCoT_MAT-SED}
\end{figure*}

\section{Applying DeepCoT to MAT-SED}\label{sc:mat-sed_DeepCoT}
\subsection{Architecture modifications}
MAT-SED \cite{Cai24MAT_SED} is an architecture designed for Sound Event Detection (SED) for self-supervised learning scenarios. It is composed by a encoder network consisting on a 2D temporal Convolution followed by a stack of 10 encoder layers \cite{Vaswani17transformer}. The outputs of this encoder are used as inputs for the clip classification head, and for the context network. This context network is composed by 3 TransformerXL \cite{Dai19TransformerXL} layers. This is then used to perform frame and clip level predictions.

From the architecture perspective, we have made the following changes to MAT-SED to transform it into an efficient, DeepCoT-based architecture:
\begin{itemize}
    \item The initial 2D temporal convolution layer is replaced by a redundancy-free Continual Convolution \cite{Hedegaard23cont_trans} to enable Continual Inference.
    \item All 10 layers in the encoder are replaced with DeepCoT layers. This affects the forward computation of both the global and local feature extractions, with separate Key and Value memories for each extraction branch. The Absolute Positional Encoding is replaced by a Recycling Positional Encoding \cite{Hedegaard23cont_trans}, and the CLS and distillation tokens are removed during inference.
    \item All three Transformer-XL encoder layers \cite{Dai19TransformerXL} are also optimized for redundancy-free DeepCoT attention.
\end{itemize}

These changes can be visualized in Fig \cref{fig:DeepCoT_MAT-SED}. During Continual Inference, the features outputted by the Continual Convolution are flattened, producing 12 tokens that are fed to the DeepCoT encoder. These are handled simultaneously for every DeepCoT layer, \ie these DeepCoT layers receive 12 tokens and output 12 tokens. Additionally, the model uses the Patch Transformer module two times: once to focus on the full sequence and another time that focuses exclusively on the last chunk. As such, separate sets of Continual Inference memories are used for each case.

\subsubsection{TransformerXL modifications}
TransformerXL proposes an attention modification to enable attention at a global context, by proposing the following attention product \cite{Dai19TransformerXL}:
\begin{equation}
    \alpha_{\text{XL}} = \text{softmax}((Q_\text{u}K^T + Q_\text{v} P)\lambda)V,
\end{equation}
where $Q_\text{u}$ is the query enhanced by a learned global token bias, $Q_\text{v}$ is the query enhanced by a learnable bias, $P$ is a positional embedding matrix of size $d \times n$, and $\lambda$ is a constant used as a scaling factor. After applying the DeepCoT modifications, the attention product is redefined as:
\begin{equation}
    \alpha_{\text{DeepCoT\_XL}} = \text{softmax}((q_\text{u}K_\text{mem}^T + q_\text{v} P)\lambda)V_\text{mem},
\end{equation}
where $K_\text{mem}$ and $V_\text{mem}$ are the memory matrices of the Key and Value, $q_\text{u}$ and $q_\text{v}$ are the enhanced tokens of the query corresponding to the newest time step. With this change we can enable efficient Continual Inference while keeping the properties of TransformerXL. This also shows how other attention mechanisms can be adapted for faster inference.

In our experiments, the DeepCoT Transformer XL modules handle 10 tokens simultaneously in this setup, analogously to the DeepCoT layers in the Patch Transformer module. 

\subsection{MAT-SED finetuning}
We have applied some modifications for the DeepCoT finetuning for Task 4 in the DCASE2023 dataset \cite{Turpault19DCASE}. We have used the checkpoint from the base model after the first finetuning stage, made the DeepCoT modifications, and then perform the second finetuning stage with DeepCoT. The motivation behind this is that the first stage only updates the classification heads, while the second stage updates the weights through the entire model.

The training of MAT-SED for the URBAN-SED \cite{Salamon17URBAN_SED} dataset uses the pretrained checkpoint from \cite{Cai24MAT_SED}, which is then finetuned for 60 epochs with the frame and clip level losses from the original model. As such, the mean-teacher model used in the original MAT-SED is not used to finetune this dataset, as all URBAN-SED data is strongly labeled. In this case, the DeepCoT-based model also gets finetuned in the same way as its non-DeepCoT counterpart.

\subsection{Availability of DCASE2023 Task 4}

Some of the strongly labeled training data of the DCASE 2023 Task 4 dataset \cite{Turpault19DCASE} is not available anymore due to broken links. In particular, only 1787/3470 clips ($\sim$ 51\%) are available, which impacts negatively the performance of the trained models.

\section{Setup for sliding window-based processing of text data}
\label{sc:text_finetuning}
For fine-tuning non-DeepCoT models, we slide an attention window of $n$ elements over every input in the training set, and use the samples corresponding to those windows for training. The samples with less than $n$ tokens are also used for training and evaluation without any modification. To compute the evaluation performance of DeepCoT models, the full input text sequence is introduced one token at a time for every sample, both at training and evaluation time, and the classification is only performed over at the last output token of the sequence to replicate a Continual Inference evaluation setup. For non-DeepCoT models, the evaluation performance is calculated by performing the classification over the last output token of the text sequences in the evaluation sets, \ie using the last window with $n$ tokens as input, in order to have performance evaluation over the same token as for the DeepCoT models.

[CLS] tokens are also removed for all models, as deeper Continual Transformers do not benefit by using them \cite{Hedegaard23cont_trans} and they introduce some overhead in Continual Inference settings. Instead, we use the output token corresponding to the newest element in the attention window.

We only use the sequences in the validation set that have at least as many elements as the attention window ($n$) to evaluate the runtime. Since MRPC dataset does not have any sequence with 104 tokens or more, there are no available runtime results for that experiment. This also implies that this particular experiment has been fine-tuned and evaluated in an identical way as previous BERT-like architectures \cite{Devlin19BERT, Su24Roformer}.

To stabilize the training of SOFT-based models the learning rate is reduced to $1e^{-6}$, 0.1 epochs of warm-up training are added, and  a maximum limit of 0.5 is set to the gradients, to ensure proper convergence.

Additionally, some DeepCoT SOFT Roformer trainings are done with smaller batch sizes due to out-of-memory errors. This is caused because the gradients for every inference step must be stored for DeepCoT-based models, increasing the memory cost during training time. In particular, MNLI and QNLI trainings with window sizes of $\times 1$ were done with a batch size of 16, QQP and MNLI trainings with window sizes of $\times 2$ were done with a batch size of 16, and QNLI training with a window size of $\times 2$ was done with a batch size of 8.

\begin{figure*}
  \centering
  \subfloat[Batch size = 1]{%
    \includegraphics[width=0.48\linewidth]{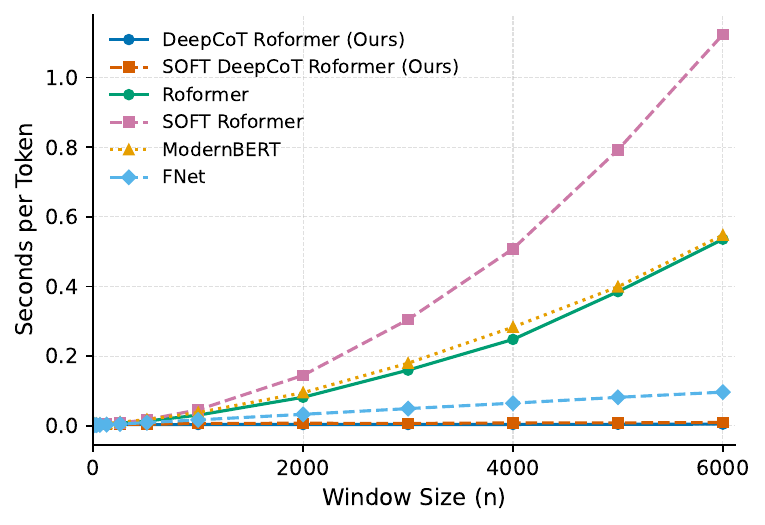}
    \label{fig:spt_1}
    }
  \hfill
  \subfloat[Batch size = 16]{%
    \includegraphics[width=0.48\linewidth]{fig/runtime_16.pdf}
    \label{fig:spt_16}
  }
  \caption{Average latency (seconds per token) observed with different window sizes.}
  \label{fig:runtime_spt}
\end{figure*}

\begin{figure*}
  \centering
  \subfloat[Batch size = 1]{%
    \includegraphics[width=0.48\linewidth]{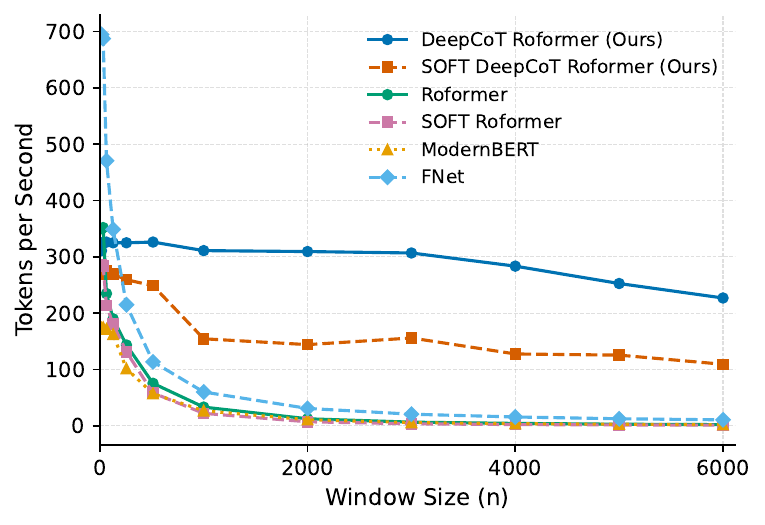}
    \label{fig:tps_1}
    }
  \hfill
  \subfloat[Batch size = 16]{%
    \includegraphics[width=0.48\linewidth]{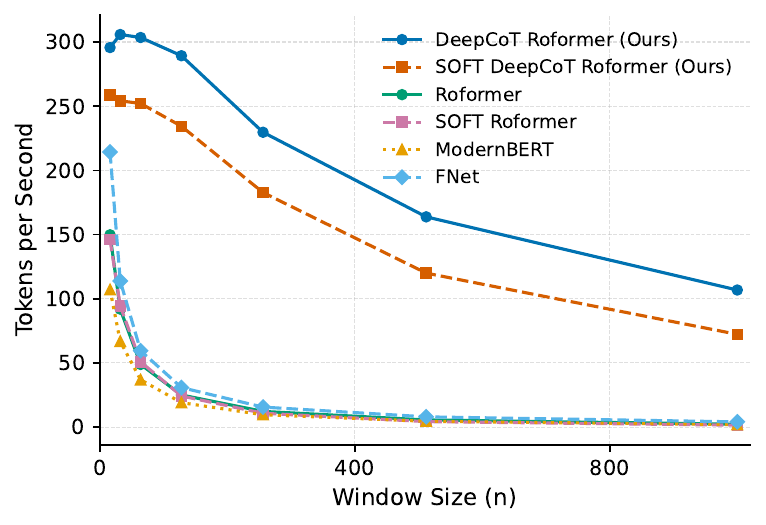}
    \label{fig:tps_16}
  }
  \caption{Average throughput (tokens per second) observed with different window sizes.}
  \label{fig:runtime_tps}
\end{figure*}

\section{Additional runtime experiments}\label{sc:runtime_supplementary}
We report additional results on the runtime of different models as a function of the input window size ($n$) in \cref{fig:runtime_spt} and \cref{fig:runtime_tps}. These results were obtained after performing inference on a single NVIDIA RTX 2080 Ti GPU. For all non-DeepCoT models, there is a sharp increase in inference time for window sizes larger than 128 tokens, associated to their higher asymptotic costs. The differences are further increased when using larger batch sizes. FNet \cite{Lee22FNet} has the highest throughput for smaller window sizes. However, its throughput falls behind to that of DeepCoT models for larger window sizes due to its higher computational cost of $O(n\text{log}(n))$. Moreover, the overhead of FNet increases with higher batch sizes more than the other models. ModernBERT \cite{Warner25ModernBERT} and Roformer \cite{Su24Roformer} show similar runtime values. However, the Roformer has a slightly smaller overhead. This difference becomes bigger with larger batch sizes.

The SOFT-based models show a higher cost compared to their non-SOFT counterparts. These differences correspond to those of a small multiplicative factor rather than an asymptotic difference.
DeepCoT models show a faster throughput due to their linear computational cost and redundant-free computations over data streams. As such, only a small decrease in throughput is observed for large window sizes. Moreover, DeepCoTs still have access to a larger effective temporal window size compared to other models, making them a good model choice for models that require long temporal contexts.

%% file: main.bib
@String(CVPR  = {IEEE Conf. Comput. Vis. Pattern Recog.})

@String(ICCV  = {Int. Conf. Comput. Vis.})

@String(ECCV  = {Eur. Conf. Comput. Vis.})

@String(ICML  = {Int. Conf. Mach. Learn.})

@String(ICLR  = {Int. Conf. Learn. Represent.})

@String(AAAI  = {AAAI})

@String(CVPR  = {CVPR})

@String(ICCV  = {ICCV})

@String(ECCV  = {ECCV})

@String(ICML  = {ICML})

@String(ICLR  = {ICLR})

@inproceedings{Hedegaard23cont_trans,
  author = {Lukas Hedegaard and Arian Bakhtiarnia and Alexandros Iosifidis},
  title = {Continual Transformers: Redundancy-Free Attention for Online Inference},
  booktitle = {International Conference on Learning Representations},
  year = {2023} 
}

@inproceedings{Vaswani17transformer,
  author       = {Ashish Vaswani and
                  Noam Shazeer and
                  Niki Parmar and
                  Jakob Uszkoreit and
                  Llion Jones and
                  Aidan N. Gomez and
                  Lukasz Kaiser and
                  Illia Polosukhin},
  title        = {Attention is All you Need},
  booktitle    = {Advances in Neural Information Processing Systems},
  year         = {2017},
}

@inproceedings{Brown20GPT3,
  author       = {Brown, Tom and Mann, Benjamin and Ryder, Nick and Subbiah, Melanie and Kaplan, Jared D and Dhariwal, Prafulla and Neelakantan, Arvind and Shyam, Pranav and Sastry, Girish and Askell, Amanda and Agarwal, Sandhini and Herbert-Voss, Ariel and Krueger, Gretchen and Henighan, Tom and Child, Rewon and Ramesh, Aditya and Ziegler, Daniel and Wu, Jeffrey and Winter, Clemens and Hesse, Chris and Chen, Mark and Sigler, Eric and Litwin, Mateusz and Gray, Scott and Chess, Benjamin and Clark, Jack and Berner, Christopher and McCandlish, Sam and Radford, Alec and Sutskever, Ilya and Amodei, Dario},
  title        = {Language Models are Few-Shot Learners},
  booktitle    = {Neural Information Processing Systems},
  year         = {2020},
}

@inproceedings{Lu21SOFT,
  author       = {Jiachen Lu and
                  Jinghan Yao and
                  Junge Zhang and
                  Xiatian Zhu and
                  Hang Xu and
                  Weiguo Gao and
                  Chunjing Xu and
                  Tao Xiang and
                  Li Zhang},
  title        = {SOFT: Softmax-free Transformer with Linear Complexity},
  booktitle    = {Advances in Neural Information Processing Systems.},
  year         = {2021},
}

@article{Ba16Layer_Normalization,
  author       = {Lei Jimmy Ba and
                  Jamie Ryan Kiros and
                  Geoffrey E. Hinton},
  title        = {Layer Normalization},
  journal      = {arXiv:1607.06450},
  volume       = {abs/1607.06450},
  year         = {2016},
}

@inproceedings{Das24TimesFM,
  author       = {Abhimanyu Das and
                  Weihao Kong and
                  Rajat Sen and
                  Yichen Zhou},
  title        = {A decoder-only foundation model for time-series forecasting},
  booktitle    = {International Conference on Machine Learning},
  year         = {2024},
}

@inproceedings{Faw25Stitching_time,
    title={In-Context Fine-Tuning for Time-Series Foundation Models},
    author={Matthew Faw and Rajat Sen and Yichen Zhou and Abhimanyu Das},
    booktitle={International Conference on Machine Learning},
    year={2025}
}

@inproceedings{Shi24KV_optimal,
    title={Keep the Cost Down: A Review on Methods to Optimize {LLM}{\textquoteright}s {KV}-Cache Consumption},
    author={Shi Luohe and Hongyi Zhang and Yao Yao and Zuchao Li and Hai Zhao},
    booktitle={Conference on Language Modeling},
    year={2024}
}

@article{Li25KV_Acceleration,
  author       = {Haoyang Li and
                  Yiming Li and
                  Anxin Tian and
                  Tianhao Tang and
                  Zhanchao Xu and
                  Xuejia Chen and
                  Nicole Hu and
                  Wei Dong and
                  Qing Li and
                  Lei Chen},
  title        = {A Survey on Large Language Model Acceleration based on {KV} Cache Management},
  journal      = {Transactions on Machine Learning Research},
  volume       = {2025},
  year         = {2025},
}

@inproceedings{Qorib24EncodervsDecoder,
  author       = {Qorib, Muhammad Reza and Moon, Geonsik and Ng, Hwee Tou},
  title        = {Are Decoder-Only Language Models Better than Encoder-Only Language Models in Understanding Word Meaning?},
  booktitle    = {Findings of the Association for Computational Linguistics, {ACL}},
  year         = {2024}
}

@inproceedings{Hedegaard22continual_inference,
  author       = {Lukas Hedegaard and Alexandros Iosifidis},
  title        = {Continual Inference: {A} Library for Efficient Online Inference with Deep Neural Networks in PyTorch},
  booktitle    = {Computer Vision - {ECCV} Workshops},
  year         = {2022},
}

@inproceedings{Wolf20hf,
  author       = {Thomas Wolf and
                  Lysandre Debut and
                  Victor Sanh and
                  Julien Chaumond and
                  Clement Delangue and
                  Anthony Moi and
                  Pierric Cistac and
                  Tim Rault and
                  R{\'{e}}mi Louf and
                  Morgan Funtowicz and
                  Joe Davison and
                  Sam Shleifer and
                  Patrick von Platen and
                  Clara Ma and
                  Yacine Jernite and
                  Julien Plu and
                  Canwen Xu and
                  Teven Le Scao and
                  Sylvain Gugger and
                  Mariama Drame and
                  Quentin Lhoest and
                  Alexander M. Rush},
  title        = {Transformers: State-of-the-Art Natural Language Processing},
  booktitle    = {Proceedings of the Conference on Empirical Methods in Natural
                  Language Processing: System Demonstrations},
  year         = {2020},
}

@article{Carreto24cont_nyst,
  title = {Continual low-rank scaled dot-product attention},
  journal = {Neural Networks},
  volume = {197},
  pages = {108517},
  author={Carreto Pic{\'o}n, Gin{\'e}s and Oleksiienko, Illia and Hedegaard, Lukas and Bakhtiarnia, Arian and Iosifidis, Alexandros},
  year         = {2026}
}

@inproceedings{Hedegaard22cont_3dconv,
  author       = {Lukas Hedegaard and Alexandros Iosifidis},
  title        = {Continual 3D Convolutional Neural Networks for Real-time Processing of Videos},
  booktitle    = {European Conference on Computer Vision},
  year         = {2022}
}

@article{Hedegaard23cont_convgraph,
  author       = {Lukas Hedegaard and
                  Negar Heidari and
                  Alexandros Iosifidis},
  title        = {Continual spatio-temporal graph convolutional networks},
  journal      = {Pattern Recognition},
  volume       = {140},
  pages        = {109528},
  year         = {2023}
}

@article{Wang20Linformer,
  author       = {Sinong Wang and
                  Belinda Z. Li and
                  Madian Khabsa and
                  Han Fang and
                  Hao Ma},
  title        = {Linformer: Self-Attention with Linear Complexity},
  journal      = {arXiv:2006.04768},
  volume       = {abs/2006.04768},
  year         = {2020},
}

@inproceedings{Devlin19BERT,
  author       = {Jacob Devlin and
                  Ming{-}Wei Chang and
                  Kenton Lee and
                  Kristina Toutanova},
  title        = {{BERT:} Pre-training of Deep Bidirectional Transformers for Language Understanding},
  booktitle    = {Conference of the North American Chapter of the Association for Computational Linguistics: Human Language Technologies},
  year         = {2019},
}

@inproceedings{Wang19GLUE,
  author       = {Alex Wang and
                  Amanpreet Singh and
                  Julian Michael and
                  Felix Hill and
                  Omer Levy and
                  Samuel R. Bowman},
  title        = {{GLUE:} {A} Multi-Task Benchmark and Analysis Platform for Natural Language Understanding},
  booktitle    = {International Conference on Learning Representations},
  year         = {2019}
}

@article{Su24Roformer,
  author       = {Jianlin Su and
                  Murtadha H. M. Ahmed and
                  Yu Lu and
                  Shengfeng Pan and
                  Wen Bo and
                  Yunfeng Liu},
  title        = {RoFormer: Enhanced transformer with Rotary Position Embedding},
  journal      = {Neurocomputing},
  volume       = {568},
  pages        = {127063},
  year         = {2024}
}

@inproceedings{Lee22FNet,
  author       = {James Lee{-}Thorp and
                  Joshua Ainslie and
                  Ilya Eckstein and
                  Santiago Onta{\~{n}}{\'{o}}n},
  title        = {FNet: Mixing Tokens with Fourier Transforms},
  booktitle    = {Proceedings of Conference of the North American Chapter of
                  the Association for Computational Linguistics: Human Language Technologies},
  year         = {2022}
}

@inproceedings{Warner25ModernBERT,
  author       = {Benjamin Warner and
                  Antoine Chaffin and
                  Benjamin Clavi{\'{e}} and
                  Orion Weller and
                  Oskar Hallstr{\"{o}}m and
                  Said Taghadouini and
                  Alexis Gallagher and
                  Raja Biswas and
                  Faisal Ladhak and
                  Tom Aarsen and
                  Griffin Thomas Adams and
                  Jeremy Howard and
                  Iacopo Poli},
  title        = {Smarter, Better, Faster, Longer: {A} Modern Bidirectional Encoder for Fast, Memory Efficient, and Long Context Finetuning and Inference},
  booktitle    = {Proceedings of the Annual Meeting of the Association for Computational
                  Linguistics, {ACL}},
  year         = {2025}
}

@inproceedings{Kim22LTP,
  author       = {Sehoon Kim and
                  Sheng Shen and
                  David Thorsley and
                  Amir Gholami and
                  Woosuk Kwon and
                  Joseph Hassoun and
                  Kurt Keutzer},
  title        = {Learned Token Pruning for Transformers},
  booktitle    = {{KDD}: The {ACM} {SIGKDD} Conference on Knowledge Discovery and Data Mining},
  year         = {2022},
}

@inproceedings{Arnab21ViViT,
  author       = {Anurag Arnab and
                  Mostafa Dehghani and
                  Georg Heigold and
                  Chen Sun and
                  Mario Lucic and
                  Cordelia Schmid},
  title        = {ViViT: {A} Video Vision Transformer},
  booktitle    = {{IEEE/CVF} International Conference on Computer Vision, {ICCV}},
  year         = {2021}
}

@inproceedings{Bertasius21TimeSformer,
  author       = {Gedas Bertasius and
                  Heng Wang and
                  Lorenzo Torresani},
  title        = {Is Space-Time Attention All You Need for Video Understanding?},
  booktitle    = {Proceedings of the International Conference on Machine Learning, {ICML}},
  year         = {2021}
}

@article{Chen25VideoTrans,
  title={Understanding Video Transformers: A Review on Key Strategies for Feature Learning and Performance Optimization},
  author={Chen, Nan and Xu, Tie and Sun, Mingrui and Yao, Chenggui and Yang, Dongping},
  journal={Intelligent Computing},
  volume={4},
  pages={0143},
  year={2025}
}

@inproceedings{Liu22VideoSwin,
  author       = {Ze Liu and
                  Jia Ning and
                  Yue Cao and
                  Yixuan Wei and
                  Zheng Zhang and
                  Stephen Lin and
                  Han Hu},
  title        = {Video Swin Transformer},
  booktitle    = {{IEEE/CVF} Conference on Computer Vision and Pattern Recognition, {CVPR}},
  year         = {2022}
}

@inproceedings{Ren22DWViT,
  author       = {Pengzhen Ren and
                  Changlin Li and
                  Guangrun Wang and
                  Yun Xiao and
                  Qing Du and
                  Xiaodan Liang and
                  Xiaojun Chang},
  title        = {Beyond Fixation: Dynamic Window Visual Transformer},
  booktitle    = {{IEEE/CVF} Conference on Computer Vision and Pattern Recognition {CVPR}},
  year         = {2022}
}

@inproceedings{Zhang22VariedWindowViT,
  author       = {Qiming Zhang and
                  Yufei Xu and
                  Jing Zhang and
                  Dacheng Tao},
  title        = {{VSA:} Learning Varied-Size Window Attention in Vision Transformers},
  booktitle    = {European Conference on Computer Vision},
  year         = {2022}
}

@inproceedings{Katharopoulos20LinearAttention,
  author       = {Angelos Katharopoulos and
                  Apoorv Vyas and
                  Nikolaos Pappas and
                  Fran{\c{c}}ois Fleuret},
  title        = {Transformers are RNNs: Fast Autoregressive Transformers with Linear Attention},
  booktitle    = {Proceedings of the International Conference on Machine Learning, {ICML}},
  year         = {2020},
}

@inproceedings{Xiong21Nystromformers,
  author       = {Yunyang Xiong and
                  Zhanpeng Zeng and
                  Rudrasis Chakraborty and
                  Mingxing Tan and
                  Glenn Fung and
                  Yin Li and
                  Vikas Singh},
  title        = {Nystr{\"{o}}mformer: {A} Nystr{\"{o}}m-based Algorithm for Approximating Self-Attention},
  booktitle    = {Proceedings of the {AAAI} Conference on Artificial Intelligence},
  year         = {2021},
}

@article{McDermott25LoLA,
  author       = {Luke McDermott and
                  Robert W. Heath Jr. and
                  Rahul Parhi},
  title        = {LoLA: Low-Rank Linear Attention With Spars Caching},
  journal      = {arXiv:2505.23666},
  volume       = {abs/2505.23666},
  year         = {2025},
}

@article{Raiaan24LLMReview,
  author       = {Mohaimenul Azam Khan Raiaan and
                  Md. Saddam Hossain Mukta and
                  Kaniz Fatema and
                  Nur Mohammad Fahad and
                  Sadman Sakib and
                  Most Marufatul Jannat Mim and
                  Jubaer Ahmad and
                  Mohammed Eunus Ali and
                  Sami Azam},
  title        = {A Review on Large Language Models: Architectures, Applications, Taxonomies, Open Issues and Challenges},
  journal      = {{IEEE} Access},
  volume       = {12},
  pages        = {26839--26874},
  year         = {2024}
}

@inproceedings{He23DeBERTaV3,
  author       = {Pengcheng He and
                  Jianfeng Gao and
                  Weizhu Chen},
  title        = {DeBERTaV3: Improving DeBERTa using ELECTRA-Style Pre-Training with Gradient-Disentangled Embedding Sharing},
  booktitle    = {International Conference on Learning Representations, {ICLR}},
  year         = {2023}
}

@inproceedings{Zhuang21RoBERTa,
    title = "A Robustly Optimized {BERT} Pre-training Approach with Post-training",
    author = "Zhuang, Liu  and
      Wayne, Lin  and
      Ya, Shi  and
      Jun, Zhao",
    booktitle = "Proceedings of the Chinese National Conference on Computational Linguistics",
    year = "2021",
}

@inproceedings{Dosovitskiy21ViT,
  author       = {Alexey Dosovitskiy and
                  Lucas Beyer and
                  Alexander Kolesnikov and
                  Dirk Weissenborn and
                  Xiaohua Zhai and
                  Thomas Unterthiner and
                  Mostafa Dehghani and
                  Matthias Minderer and
                  Georg Heigold and
                  Sylvain Gelly and
                  Jakob Uszkoreit and
                  Neil Houlsby},
  title        = {An Image is Worth 16x16 Words: Transformers for Image Recognition at Scale},
  booktitle    = {International Conference on Learning Representations, {ICLR}},
  year         = {2021},
}

@inproceedings{Faw25FewShotTimesFM,
title={In-Context Fine-Tuning for Time-Series Foundation Models},
author={Matthew Faw and Rajat Sen and Yichen Zhou and Abhimanyu Das},
booktitle={International Conference on Machine Learning},
year={2025},
}

@article{Xiao25MiniCPM,
  author       = {Chaojun Xiao and
                  Yuxuan Li and
                  Xu Han and
                  Yuzhuo Bai and
                  Jie Cai and
                  Haotian Chen and
                  Wentong Chen and
                  Xin Cong and
                  Ganqu Cui and
                  Ning Ding and
                  Shengda Fan and
                  Yewei Fang and
                  Zixuan Fu and
                  Wenyu Guan and
                  Yitong Guan and
                  Junshao Guo and
                  Yufeng Han and
                  Bingxiang He and
                  Yuxiang Huang and
                  Cunliang Kong and
                  Qiuzuo Li and
                  Siyuan Li and
                  Wenhao Li and
                  Yanghao Li and
                  Yishan Li and
                  Zhen Li and
                  Dan Liu and
                  Biyuan Lin and
                  Yankai Lin and
                  Xiang Long and
                  Quanyu Lu and
                  Yaxi Lu and
                  Peiyan Luo and
                  Hongya Lyu and
                  Litu Ou and
                  Yinxu Pan and
                  Zekai Qu and
                  Qundong Shi and
                  Zijun Song and
                  Jiayuan Su and
                  Zhou Su and
                  Ao Sun and
                  Xianghui Sun and
                  Peijun Tang and
                  Fangzheng Wang and
                  Feng Wang and
                  Shuo Wang and
                  Yudong Wang and
                  Yesai Wu and
                  Zhenyu Xiao and
                  Jie Xie and
                  Zihao Xie and
                  Yukun Yan and
                  Jiarui Yuan and
                  Kaihuo Zhang and
                  Lei Zhang and
                  Linyue Zhang and
                  Xueren Zhang and
                  Yudi Zhang and
                  Hengyu Zhao and
                  Weilin Zhao and
                  Weilun Zhao and
                  Yuanqian Zhao and
                  Zhi Zheng and
                  Ge Zhou and
                  Jie Zhou and
                  Wei Zhou and
                  Zihan Zhou and
                  Zixuan Zhou and
                  Zhiyuan Liu and
                  Guoyang Zeng and
                  Chao Jia and
                  Dahai Li and
                  Maosong Sun},
  title        = {MiniCPM4: Ultra-Efficient LLMs on End Devices},
  journal      = {arXiv:2506.07900},
  volume       = {abs/2506.07900},
  year         = {2025},
}

@inproceedings{He21Deberta,
  author       = {Pengcheng He and
                  Xiaodong Liu and
                  Jianfeng Gao and
                  Weizhu Chen},
  title        = {Deberta: decoding-Enhanced Bert with Disentangled Attention},
  booktitle    = {International Conference on Learning Representations, {ICLR}},
  year         = {2021},
}

@inproceedings{Obeidat25EncoderBiomedical,
  author       = {Motasem S. Obeidat and
                  Md Sultan Al Nahian and
                  Ramakanth Kavuluru},
  title        = {Do LLMs Surpass Encoders for Biomedical NER?},
  booktitle    = {{IEEE} International Conference on Healthcare Informatics, {ICHI}},
  year         = {2025}
}

@article{Child19SparseTransformers,
  author       = {Rewon Child and
                  Scott Gray and
                  Alec Radford and
                  Ilya Sutskever},
  title        = {Generating Long Sequences with Sparse Transformers},
  journal      = {arXiv:1904.10509},
  volume       = {abs/1904.10509},
  year         = {2019},
}

@article{Beltagy19Longformer,
  author       = {Iz Beltagy and
                  Matthew E. Peters and
                  Arman Cohan},
  title        = {Longformer: The Long-Document Transformer},
  journal      = {arXiv:2004.05150},
  volume       = {abs/2004.05150},
  year         = {2020}
}

@article{Jiang23Mistral,
      title={Mistral 7B}, 
      author={Albert Q. Jiang and Alexandre Sablayrolles and Arthur Mensch and Chris Bamford and Devendra Singh Chaplot and Diego de las Casas and Florian Bressand and Gianna Lengyel and Guillaume Lample and Lucile Saulnier and Lélio Renard Lavaud and Marie-Anne Lachaux and Pierre Stock and Teven Le Scao and Thibaut Lavril and Thomas Wang and Timothée Lacroix and William El Sayed},
      year={2023},
      journal={arXiv:2310.06825},
      volume={abs/2310.06825},
}

@inproceedings{Dao22FlashAttention,
  author       = {Tri Dao and
                  Daniel Y. Fu and
                  Stefano Ermon and
                  Atri Rudra and
                  Christopher R{\'{e}}},
  title        = {FlashAttention: Fast and Memory-Efficient Exact Attention with IO-Awareness},
  booktitle    = {Advances in Neural Information Processing Systems},
  year         = {2022}
}

@inproceedings{Shah24FlashAttention3,
  author       = {Jay Shah and
                  Ganesh Bikshandi and
                  Ying Zhang and
                  Vijay Thakkar and
                  Pradeep Ramani and
                  Tri Dao},
  title        = {FlashAttention-3: Fast and Accurate Attention with Asynchrony and Low-precision},
  booktitle    = {Advances in Neural Information Processing Systems},
  year         = {2024}
}

@inproceedings{Szegedy16Inception,
  author       = {Christian Szegedy and
                  Vincent Vanhoucke and
                  Sergey Ioffe and
                  Jonathon Shlens and
                  Zbigniew Wojna},
  title        = {Rethinking the Inception Architecture for Computer Vision},
  booktitle    = {{IEEE} Conference on Computer Vision and Pattern Recognition, {CVPR}},
  year         = {2016}
}

@article{Sun25EfficientLLMSurvey,
  author       = {Weigao Sun and
                  Jiaxi Hu and
                  Yucheng Zhou and
                  Jusen Du and
                  Disen Lan and
                  Kexin Wang and
                  Tong Zhu and
                  Xiaoye Qu and
                  Yu Zhang and
                  Xiaoyu Mo and
                  Daizong Liu and
                  Yuxuan Liang and
                  Wenliang Chen and
                  Guoqi Li and
                  Yu Cheng},
  title        = {Speed Always Wins: {A} Survey on Efficient Architectures for Large Language Models},
  journal      = {arXiv:2508.09834},
  volume       = {abs/2508.09834},
  year         = {2025}
}

@article{Zaman25audiotransformers,
  author       = {Khalid Zaman and
                  Kai Li and
                  Melike Sah and
                  Cem Direkoglu and
                  Shogo Okada and
                  Masashi Unoki},
  title        = {Transformers and audio detection tasks: An overview},
  journal      = {Digital Signal Processing},
  volume       = {158},
  pages        = {104956},
  year         = {2025}
}

@article{Tzanetakis02GTZAN,
  author       = {George Tzanetakis and
                  Perry R. Cook},
  title        = {Musical genre classification of audio signals},
  journal      = {IEEE Transactions on speech and audio processing},
  volume       = {10},
  number       = {5},
  pages        = {293--302},
  year         = {2002}
}

@inproceedings{williams18MNLI,
  title={A broad-coverage challenge corpus for sentence understanding through inference},
  author={Williams, Adina and Nangia, Nikita and Bowman, Samuel},
  booktitle={Proceedings conference of the North American chapter of the association for computational linguistics},
  year={2018}
}

@inproceedings{DeGeest16OAD,
  author       = {Roeland De Geest and
                  Efstratios Gavves and
                  Amir Ghodrati and
                  Zhenyang Li and
                  Cees Snoek and
                  Tinne Tuytelaars},
  title        = {Online Action Detection},
  booktitle    = {European Conference on Computer Vision},
  year         = {2016},
}

@inproceedings{Pavlopoulos20ClinicalKeyboard,
  title={Clinical predictive keyboard using statistical and neural language modeling},
  author={Pavlopoulos, John and Papapetrou, Panagiotis},
  booktitle={International Symposium on Computer-Based Medical Systems (CBMS)},
  year={2020},
}

@inproceedings{Wang21Oadtr,
  title={Oadtr: Online action detection with transformers},
  author={Wang, Xiang and Zhang, Shiwei and Qing, Zhiwu and Shao, Yuanjie and Zuo, Zhengrong and Gao, Changxin and Sang, Nong},
  booktitle={Proceedings of the IEEE/CVF International Conference on Computer Vision},
  year={2021}
}

@article{Idrees17THUMOS,
  author       = {Haroon Idrees and
                  Amir R. Zamir and
                  Yu{-}Gang Jiang and
                  Alex Gorban and
                  Ivan Laptev and
                  Rahul Sukthankar and
                  Mubarak Shah},
  title        = {The {THUMOS} challenge on action recognition for videos "in the wild"},
  journal      = {Computer Vision and Image Understanding},
  volume       = {155},
  pages        = {1--23},
  year         = {2017}
}

@article{Wang19temporal_segment_networks,
  author       = {Limin Wang and
                  Yuanjun Xiong and
                  Zhe Wang and
                  Yu Qiao and
                  Dahua Lin and
                  Xiaoou Tang and
                  Luc Van Gool},
  title        = {Temporal Segment Networks for Action Recognition in Videos},
  journal      = {{IEEE} Transactions on Pattern Analysis and Machine Intelligence},
  volume       = {41},
  number       = {11},
  pages        = {2740--2755},
  year         = {2019}
}

@inproceedings{Heilbron15ActivityNet,
  author       = {Fabian Caba Heilbron and
                  Victor Escorcia and
                  Bernard Ghanem and
                  Juan Carlos Niebles},
  title        = {ActivityNet: {A} large-scale video benchmark for human activity understanding},
  booktitle    = {{IEEE} Conference on Computer Vision and Pattern Recognition},
  year         = {2015},
}

@inproceedings{Carreira17Kinetics,
  author       = {Jo{\~{a}}o Carreira and
                  Andrew Zisserman},
  title        = {Quo Vadis, Action Recognition? {A} New Model and the Kinetics Dataset},
  booktitle    = {{IEEE} Conference on Computer Vision and Pattern Recognition},
  year         = {2017}
}

@article{Slama25CoGraphTransformers,
  author       = {Rim Slama and
                  Wael Rabah and
                  Hazem Wannous},
  title        = {Online hand gesture recognition using Continual Graph Transformers},
  journal      = {arXiv:2502.14939},
  volume       = {abs/2502.14939},
  year         = {2025}
}

@inproceedings{Bachlechner21ReZero,
  author       = {Thomas Bachlechner and
                  Bodhisattwa Prasad Majumder and
                  Huanru Henry Mao and
                  Gary Cottrell and
                  Julian J. McAuley},
  title        = {ReZero is all you need: fast convergence at large depth},
  booktitle    = {Proceedings of the Conference on Uncertainty in Artificial Intelligence, {UAI}},
  year         = {2021}
}

@article{Tay22EfficientTransformersSurvey,
author = {Tay, Yi and Dehghani, Mostafa and Bahri, Dara and Metzler, Donald},
title = {Efficient Transformers: A Survey},
year = {2022},
volume = {55},
number = {6},
journal = {ACM Computing Surveys},
pages={1--28},
}

@inproceedings{Luo16ReceptiveFieldCNNs,
  author       = {Wenjie Luo and
                  Yujia Li and
                  Raquel Urtasun and
                  Richard S. Zemel},
  title        = {Understanding the Effective Receptive Field in Deep Convolutional Neural Networks},
  booktitle    = {Advances in Neural Information Processing Systems},
  year         = {2016}
}

@inproceedings{Turpault19DCASE,
    Author = "Turpault, Nicolas and Serizel, Romain and Parag Shah, Ankit and Salamon, Justin",
    title = "{Sound event detection in domestic environments with weakly labeled data and soundscape synthesis}",
    booktitle = "{Workshop on Detection and Classification of Acoustic Scenes and Events}",
    year = "2019"
}

@inproceedings{Cai24MAT_SED,
  author       = {Pengfei Cai and
                  Yan Song and
                  Kang Li and
                  Haoyu Song and
                  Ian McLoughlin},
  title        = {MAT-SED: A Masked Audio Transformer with Masked-Reconstruction Based Pre-training for Sound Event Detection},
  booktitle    = {Annual Conference of the International Speech Communication Association, Interspeech},
  year         = {2024}
}

@inproceedings{Dai19TransformerXL,
  title={Transformer-xl: Attentive language models beyond a fixed-length context},
  author={Dai, Zihang and Yang, Zhilin and Yang, Yiming and Carbonell, Jaime G and Le, Quoc and Salakhutdinov, Ruslan},
  booktitle={Proceedings annual meeting of the association for computational linguistics},
  year={2019}
}

@INPROCEEDINGS{Salamon17URBAN_SED,
  author={Salamon, Justin and MacConnell, Duncan and Cartwright, Mark and Li, Peter and Bello, Juan Pablo},
  booktitle={IEEE Workshop on Applications of Signal Processing to Audio and Acoustics (WASPAA)}, 
  title={Scaper: A library for soundscape synthesis and augmentation}, 
  year={2017},
}
